\documentclass[journal]{IEEEtran}

\ifCLASSINFOpdf
\else
\fi
\usepackage{graphicx}
\usepackage{amsmath}
\usepackage{cite}

 \newcommand{\etal}{\textit{et al}.}
\newcommand{\ie}{\textit{i}.\textit{e}.}
\newcommand{\eg}{\textit{e}.\textit{g}.}

\begin{document}

\title{A Cascaded Convolutional Neural Network for Single Image Dehazing}

\author{Chongyi~Li,~\IEEEmembership{Student~Member,~IEEE,}
        Jichang~Guo,
        Fatih~Porikli,~\IEEEmembership{Fellow,~IEEE,}
        Huazhu~Fu,
        and Yanwei~Pang~\IEEEmembership{Senior~Member,~IEEE}
\thanks{This work was supported in part by the National Key Basic Research Program of China (2014CB340403), the Natural Science Foundation of Tianjin of China (15JCYBJC15500), the National Natural Science Foundation of China (61771334), the Tianjin Research Program of Application Foundation and Advanced Technology (15JCQNJC01800), and the program of China Scholarships Council (CSC) under the Grant CSC No. 201606250063.}
\thanks{Chongyi Li is with the School of Electrical and Information Engineering, Tianjin University, Tianjin, China and Research School of Engineering, College of Engineering and Computer Science, Australian National University, Canberra, ACT 0200, Australia (e-mail: lichongyi@tju.edu.cn).

Fatih Porikli is with the Australian National University, Canberra, ACT 0200, Australia (e-mail: fatih.porikli@gmail.com).

Jichang Guo and Yanwei Pang are with the School of Electrical and Information Engineering, Tianjin University, Tianjin, China (e-mail: jcguo@tju.edu.cn; pyw@tju.edu.cn).

Huazhu Fu is with Institute for Infocomm Research, at Agency for Science, Technology and Research, Singapore (e-mail: huazhufu@gmail.com).

(Corresponding author: Jichang Guo.)}}
\markboth{IEEE ACCESS}%
{Shell \MakeLowercase{\textit{et al.}}: Bare Demo of IEEEtran.cls for Journals}

\maketitle

\begin{abstract}
Images captured under outdoor scenes usually suffer from low contrast and limited visibility due to suspended atmospheric particles, which directly affects the quality of photos. Despite numerous image dehazing methods have been proposed, effective hazy image restoration remains a challenging problem. Existing learning-based methods usually predict the medium transmission by Convolutional Neural Networks (CNNs), but ignore the key global atmospheric light. Different from previous learning-based methods, we propose a flexible cascaded CNN for single hazy image restoration, which considers the medium transmission and global atmospheric light jointly by two task-driven subnetworks. Specifically, the medium transmission estimation subnetwork is inspired by  the densely connected CNN while the global atmospheric light estimation subnetwork is a light-weight CNN. Besides, these two subnetworks are cascaded by sharing the common features. Finally, with the estimated model parameters, the haze-free image is obtained by the atmospheric scattering model inversion, which achieves more accurate and effective restoration performance. Qualitatively and quantitatively experimental results on the synthetic and real-world hazy images demonstrate that the proposed method effectively removes haze from such images, and outperforms several state-of-the-art dehazing methods.

\end{abstract}

\begin{IEEEkeywords}
Image dehazing, image degradation, image restoration, convolutional neural networks.
\end{IEEEkeywords}

\IEEEpeerreviewmaketitle

\section{Introduction}

\begin{figure}[!t]
\centering
\centerline{\includegraphics[width=8.5cm,height=2cm]{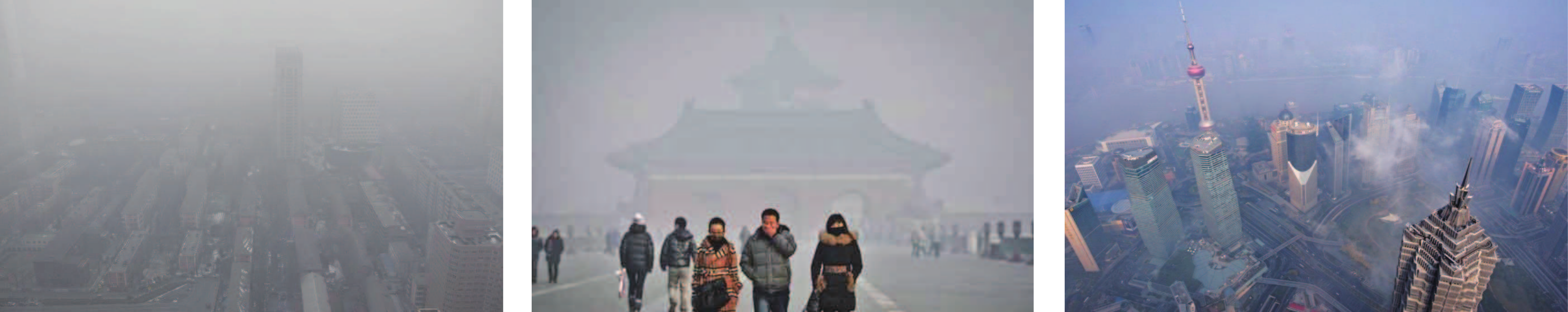}}
\caption{Several examples of images taken under hazy or foggy scenes.}
\label{fig_1}
\end{figure}

\IEEEPARstart{D}{uring} recent years, we have witnessed a rapid development of wireless network technologies and mobile devices equipped with various cameras which have revolutionized the way people take and share multimedia content \cite{Yin2014, Cong2018}. However, outdoor images (\eg, Figure~\ref{fig_1}) often suffer from low contrast, obscured clarity, and faded colors due to the floating particles in the atmosphere, such as haze, fog, or dust, that absorb and scatter light. These degraded outdoor images not only affect the quality of photos \cite{Tian2015} but also limit the applications in urban transportation \cite{Huang2014}, video analysis \cite{Zhang2012}, visual surveillance \cite{Tian2014}, and driving assistance \cite{Negru2015}. Therefore, image dehazing or image defogging has become a promising research area. Additionally, image dehazing methods also provide reference values for the underwater image enhancement and restoration research field~\cite{Li2016under,Li2017under}. However, it is still a challenging task since the haze concentration is difficult to estimate from the unknown depth in the single image.

Single hazy image restoration methods usually need to estimate two key components in the hazy image formation model (\ie, medium transmission and global atmospheric light). To achieve these two components, traditional prior-based methods either try to find new kinds of haze related priors or propose new ways to use them. However, haze related priors do not always hold, especially for the varying scenes. By contrast, to obtain more robust and accurate estimation, the learning-based methods explore the relations between the hazy images and the corresponding medium transmission in data-driven manner. However, most of the learning-based methods estimate the medium transmission and global atmospheric light separately, and do not consider the joint relations of them.
In addition, separate estimation for the medium transmission and global atmospheric light limits the flexibility of previous methods. Thus, it inspires us to explore the joint relations between the medium transmission and the global atmospheric light, and how to directly map an input hazy image to its medium transmission and global atmospheric light simultaneously in pure data-driven manner.

\textbf{Our contributions} In this paper, we propose a cascaded CNN deep model for single image dehazing.
Different from previous prior-based methods, we explore the relations between the input hazy images and the corresponding medium transmission in data-driven manner, which achieves more accurate and robust medium transmission. Compared to previous learning-based methods, we estimate the medium transmission and global atmospheric light jointly in a cascaded CNN deep model, which advances in dehazing performance and flexibility.  Additionally, compared with the existing single image dehazing methods, the proposed method has superior dehazing performance both on perceptually and quantitatively.

The rest of this paper is organized as follows: Section II presents the related work. Section III describes the proposed method in detail. Section IV presents the experimental settings, investigates the network parameter settings, and gives the experimental results. Lastly, Section V concludes and discusses this paper.

\section{Related Work}

Numerous image dehazing methods have been proposed in the recent decade \cite{Li2016survey}. These methods can be roughly classified into four categories: extra information-based methods \cite{Schechner2001,Narasimhan2003, Caraffa2012, Li2015accv}, contrast enhancement-based methods \cite{Stark2000, Kim2001, Tan2008, Ancuti2013}, prior-based methods \cite{Fattal2008, He2011, Meng2013, Fattal2014, Zhu2015, Lai2015, Baig2016, Berman2016, Chen2016, Wang2017}, and learning-based methods \cite{Tang2014, Cai2016, Ren2016, Fan2016,Li2017allinone}. Though extra information-based methods can achieve impressive dehazing performance, they show limitations in real-life applications. In general, contrast enhancement-based methods produce under or over enhanced regions, color distortion, and artifacts due to failing to consider the formation principle of the hazy image and image degradation mechanism.  As follows, we mainly introduce the prior-based and learning-based methods and summarize the existing problems.

Prior-based methods formulate some restrictions on the visual characteristics of hazy images to solve an ill-posed problem, which has made significant progress recently. Dark channel prior (DCP) method proposed by He \etal \cite {He2011} is one of classical prior-based methods, which is based on statistics that at least one channel has some pixels with very low intensities in most of non-haze patches. Based on the DCP, the medium transmission and global atmospheric light are roughly estimated. Finally, the dehazed image is achieved by the estimated medium transmission refined by soft matting \cite{Levin2006} or guided filter \cite{He2013} as well as the estimated global atmospheric light according to an atmospheric scattering model. Although, the DCP method can obtain outstanding dehazing results in most cases, it tends to over-estimate the thickness of haze, which leads to color casts, especially for the sky regions. Subsequently, many strategies are applied to enhance the performance of the original DCP method. Zhu~\etal~\cite{Zhu2015} proposed a simple yet effective prior (\ie, CAP) for image dehazing. The scene depth from the camera to the object of a hazy image is modeled in a linear model based on the CAP where unknown model parameters are estimated by a supervised learning strategy. Even though prior-based methods have achieved remarkable progress, they still have some limitations and need to be further improved. For instance, their performance is highly contingent on the accuracy of the estimated medium transmission and global atmospheric light, which is difficult to achieve when the priors are invalid. In addition, they also may entail high computation cost, which makes it infeasible for real-time applications.

With rapid development of learning technology in computer vision tasks \cite{ LeCun2015, He2015}, the learning-based methods have been adopted in image dehazing. For example, Tang \etal~\cite{Tang2014} extracted multi-scale handcrafted haze-relevant features, and then employed random forests regressor \cite{Breiman2001} to learn the correlation between the handcrafted features and the medium transmission. However, these handcrafted features are less effective and insufficient for some challenging scenes, which limits its performance. Generally, for the handcrafted features-based methods, inappropriate feature extraction often leads to poor dehazing results. Different from the handcrafted features, Cai \etal~ \cite{Cai2016} proposed a CNN-based image dehazing method, named DehazeNet, which trained a regressor to predict the medium transmission. The DehazeNet includes four sequential operations, \ie, feature extraction, multi-scale mapping, local extremum, and non-linear regression. The training dataset is generated by haze-free patches collected from Internet, random medium transmission value, and fixed global atmospheric light value (\ie, 1) based on an atmospheric scattering model. With the optimized network weights, the medium transmission of an input hazy image can be estimated by network forward propagation. After that, the guided filtering \cite{He2013} as post-processing is used to remove the blocking artifacts of the estimated medium transmission caused by the patch based estimation. Additionally, the authors applied an empirical method to estimate the global atmospheric light. Similar with DehazeNet~\cite{Cai2016}, Ren \etal~\cite{Ren2016} designed a multi-scale CNN for single image dehazing. Recently, Li \etal~\cite{Li2017allinone} proposed an all-in-one deep model for single image dehazing, which directly generated the clean image using CNN. Additionally, such all-in-one network architecture has been extended to the video dehazing~\cite{Li2017video}, which fills in the blank of video dehazing by deep learning strategies. For CNN-based methods, the accuracy of the estimated medium transmission and the dehazing performance need to be further improved, especially for varying scenes. Moreover, most of CNN-based methods estimate the global atmospheric light by the empirical methods, which limits the flexibility of network and the accuracy of restoration.

\section{Proposed Dehazing Method}

To have a better understanding of our work, we first briefly review the atmospheric scattering model and then a detailed introduction of our cascaded CNN framework and the loss functions used in the optimization is presented. Lastly, we illustrate how to use the estimated medium transmission and global atmospheric light to achieve the haze-free image. More details are introduced as follows.

\subsection{Atmospheric Scattering Model}
Haze results from air pollution such as dust, smoke, and other dry particles that obscure the clarity of sky. Image captured under hazy or foggy day, only a part of the scene reflected light reaches the imaging equipment due to the effects of atmosphere absorption and scattering caused by haze, which decreases the visibility of scene, introduces the faded colors, and reduces the visual quality.

According to the atmospheric scattering model \cite{Koschmieder1924}, a hazy image formation can be described as
 \begin{equation}
\label{equ_1}
I(x)=J(x)t(x)+B(x)(1-t(x)),
\end{equation}
where $x$ denotes a pixel, $I(x)$ is the observed image, $J(x)$ is the haze-free image, $B(x)$ is the global atmospheric light, and $t(x)\in [0,1]$ is the medium transmission which represents the percentage of the scene radiance reaching the camera. The medium transmission $t(x)$ can be further expressed in an exponential decay term as
\begin{equation}
\label{equ_2}
t(x)=\exp(-\beta d(x)).
\end{equation}
where $\beta$ is the attenuation coefficient of the atmosphere and $d(x)$ is the distance from the scene to the camera. The purpose of single image dehazing is to restore $J(x)$, $B(x)$, and $t(x)$ from $I(x)$, which is an ill-posed problem.
\subsection{The Proposed Cascaded CNN Framework}

\begin{figure*}[!htb]
\centering
\includegraphics[width=18cm,height=5cm]{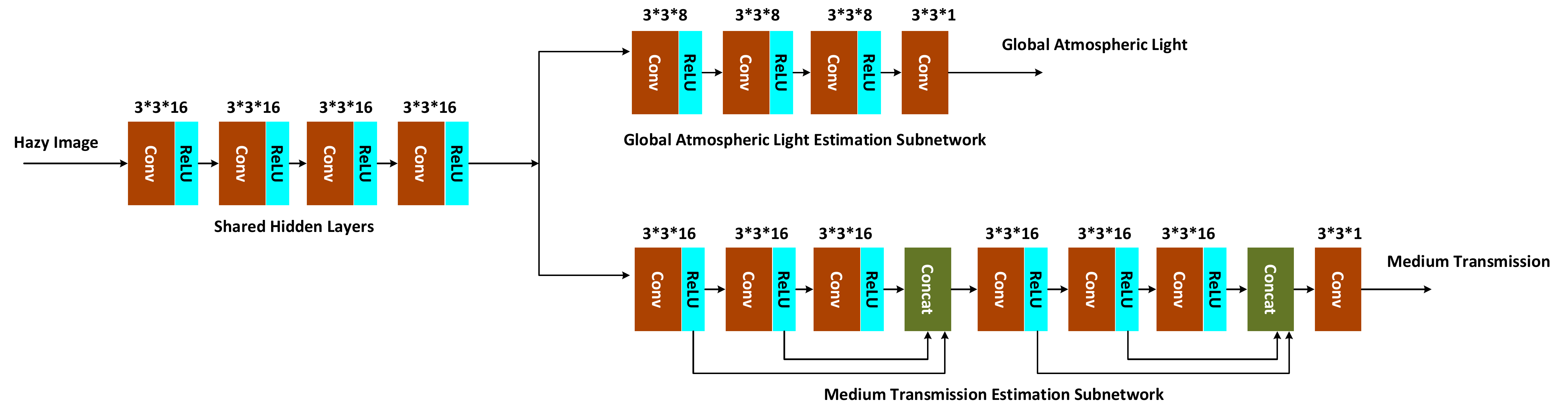}
\caption{The diagram of the proposed cascaded CNN structure. The cascaded CNN includes three parts: the shared hidden layers part, the global atmospheric light estimation subnetwork, and the medium transmission estimation subnetwork. In the network diagram, different color blocks represent the different operations. Brown block ``Conv'': convolution; light blue block ``ReLU'': ReLU nonlinearity function; dark green block ``Concat'': concatenation.}
\label{fig_6}
\end{figure*}

We aim to learn a cascaded CNN model that discerns the statistical relations between the hazy image and the corresponding medium transmission and global atmospheric light. The specific design of our cascaded CNN is presented in Figure~\ref{fig_6} for a clear explanation.

In Figure~\ref{fig_6}, the cascaded CNN includes three parts that one is the shared hidden layers part, which extracts  common features for subsequent subnetworks; one is the global atmospheric light estimation subnetwork, which takes the outputs of the shared hidden layers part as the inputs to map the global atmospheric light; one is medium transmission estimation subnetwork, which takes the outputs of the shared hidden layers part as the inputs to map the medium transmission. By such network architecture, our cascaded CNN can predict the global atmospheric light and medium transmission simultaneously.

The shared hidden layers part includes 4 convolutional layers with filter size of $f_{i}\times f_{i}\times n_{i}=3 \times 3 \times 16$ followed by ReLU nonlinearity function \cite{Krizhevsky2012}. Here, $f_{i}$ is the spatial support of a filter and $n_{i}$ is the number of filters. Since we found that the task of the global atmospheric light estimation is easy for CNN, we employ a light-weight CNN architecture for the global atmospheric light estimation subnetwork. Specifically, the global atmospheric light estimation subnetwork includes 4 convolutional layers with filter size of $ 3 \times 3 \times 8$ followed by ReLU nonlinearity function \cite{Krizhevsky2012}, except for the last one. The medium transmission estimation subnetwork architecture is inspired by the densely connected network \cite{Huang2017} which stacks early layers at the end of each block, which strengthens feature propagation and alleviates the vanishing-gradient problem. Specifically, the medium transmission estimation subnetwork includes 7 convolutional layers with filter size of $ 3 \times 3 \times 16$ followed by ReLU nonlinearity function \cite{Krizhevsky2012}, except for the last one. The network parameter settings will be discussed in Section IV. Next, we describe loss functions used in the cascaded CNN optimization.

\subsection{Loss Functions}
For image dehazing problem, most of learning-based methods employ Mean Squared Error (MSE) loss function for network optimization. Following previous methods, we also use MSE loss function for our medium transmission estimation subnetwork. For the convenience of training, we first assume that the format of the global atmospheric light is a map with dimension of $M$. Moreover, every pixel in the global atmospheric light map has the same value. Such assumption is reasonable because previous methods usually assume that every pixel in the input hazy image has the same global atmospheric light value. Then, for the global atmospheric light estimation subnetwork, we first tried MSE loss function, however, we found that the predicted global atmospheric light map is inconsistent with our assumption that every pixel in the input hazy image has the same global atmospheric light value. Thus, to avoid this problem, we use Structural Similarity Index (SSIM) loss function \cite{Zhao2017} for our global atmospheric light estimation subnetwork, which makes the values in the predicted global atmospheric light map same.

For the global atmospheric light estimation subnetwork, we minimize the SSIM loss function between the estimated global atmospheric light and the global atmospheric light ground truth. Firstly, the SSIM value for every pixel between the predicted global atmospheric light $F_{gal}(I_{i})$ and the  corresponding ground truth of the global atmospheric light $B_i$ is calculated as follows:
\begin{equation}
\label{equ_66}
SSIM(p)=\frac{2\mu_{x}\mu_{y}+C_{1}}{\mu_{x}^{2}+\mu_{y}^{2}+C_{1}}\cdot\frac{2\sigma_{xy}+C_{2}}{\sigma_{x}^{2}+\sigma_{y}^{2}+C_{2}},
\end{equation}
where $x$ and $y$ are the corresponding image patches with size $13\times13$ (default in the SSIM loss function \cite{Zhao2017}) in the predicted global atmospheric light and the corresponding ground truth, respectively. Above, $p$ is the center pixel of image patch, $\mu_{x}$ is the mean of $x$, $\sigma_{x}$ is the standard deviations of $x$, $\mu_{y}$ is the mean of $y$, $\sigma_{y}$ is the standard deviations of $y$,  $\sigma_{xy}$ is the covariance between $x$ and $y$. Using the defaults in the SSIM loss function \cite{Zhao2017}, we set the values of $C_{1}$ and $C_{2}$ to 0.02 and 0.03. In fact, our network is insensitive to those parameters. Besides, $F_{gal}$ is the learned global atmospheric light mapping function. $I_{i}$ is the input hazy image. Using Equation~\eqref{equ_66}, the SSIM loss between the predicted global atmospheric light $F_{gal}(I_{i})$ and the corresponding ground truth  $B_i$ is expressed as
\begin{equation}
\label{equ_77}
L_{{SSIM}}=\frac{1}{N}\sum_{i=1}^{N}(1-\frac{1}{M}\Sigma_{p=1}^{M}(SSIM(p))),
\end{equation}
where  $N$ is the number of each batch, $M=H\times W$ is the dimension of the predicted global atmospheric light.

For the medium transmission estimation subnetwork, we minimize the MSE loss function between the predicted medium transmission $F_{mt}(I_{i})$ and the corresponding ground truth of the medium transmission $t_i$, and is expressed as
\begin{equation}
\label{equ_5}
L_{{MSE}}=\frac{1}{NHW}\sum_{i=1}^{N}\|F_{mt}(I_{i})-t_i\|^{2},
\end{equation}
where $N$ is the number of each batch, $F_{mt}$ is the learned medium transmission mapping function, $H\times W$ is the dimension of the predicted medium transmission.

The final loss function for the cascaded CNN is the linear combination of the above-introduced losses with the following weights:
\begin{equation}
\label{equ_8}
\begin{aligned}
L_{{total}}=L_{{SSIM}}+L_{{MSE}}.
\end{aligned}
\end{equation}
The blending weights are picked empirically based on preliminary experiments on the training data, which makes the contributions of SSIM loss and MSE loss same. In addition, these two subnetworks share the weights of the shared hidden layers part and are optimized jointly.
\subsection{Haze Removal}
Finally, with the achieved medium transmission and global atmospheric light, the haze-free image can be obtained by
\begin{equation}
\label{equ_55}
J(x)=\frac{I(x)-B(x)}{t(x)}+B(x).
\end{equation}
where $J(x)$ is the haze-free image,  $I(x)$ is the input hazy image,  $B(x)$ is the estimated atmospheric light, and $t(x)$ is the estimated medium transmission refined by the guided image filtering \cite{He2013}. For our results shown in this paper, the filter size of the guided image filtering is $33\times33$.

\begin{figure*}[htb]
  \centering
\begin{minipage}[b]{0.18\linewidth}
  \centering
  \centerline{\includegraphics[width=2.9cm,height=5.7cm]{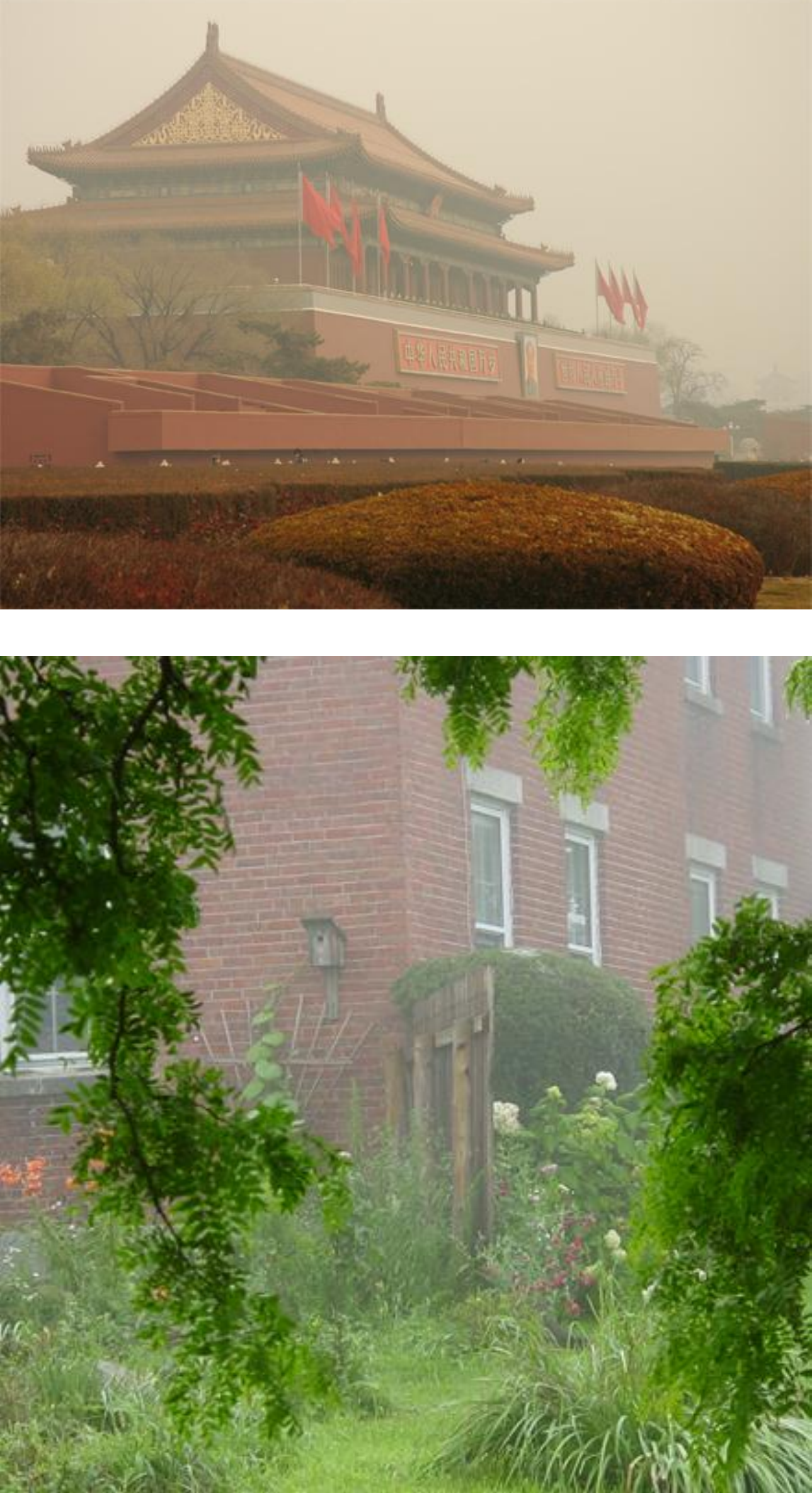}}
  \centerline{(a)}\medskip
\end{minipage}
\begin{minipage}[b]{0.18\linewidth}
  \centering
  \centerline{\includegraphics[width=2.95cm,height=5.7cm]{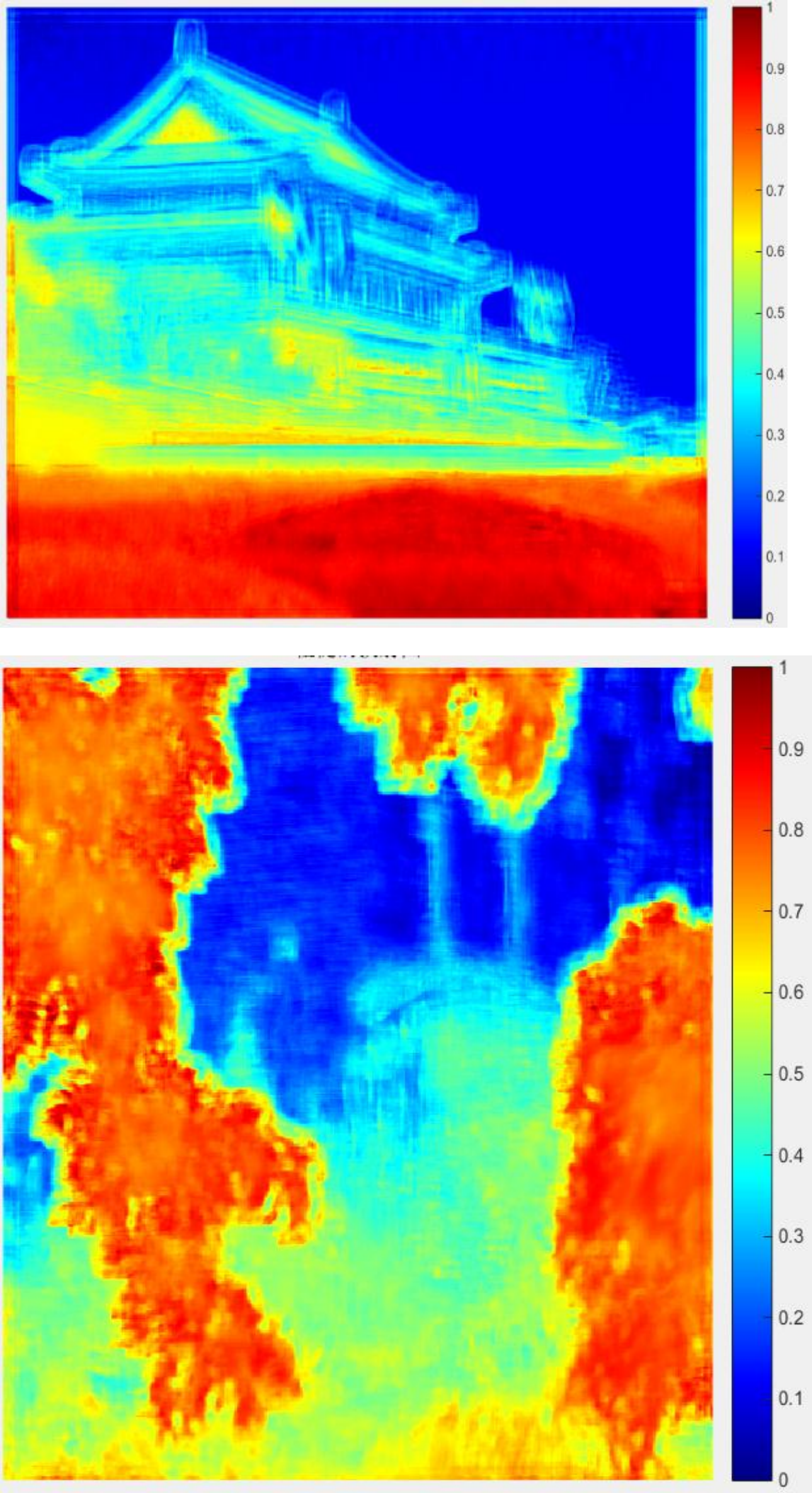}}
   \centerline{(b)}\medskip
\end{minipage}
\begin{minipage}[b]{0.18\linewidth}
  \centering
  \centerline{\includegraphics[width=2.95cm,height=5.7cm]{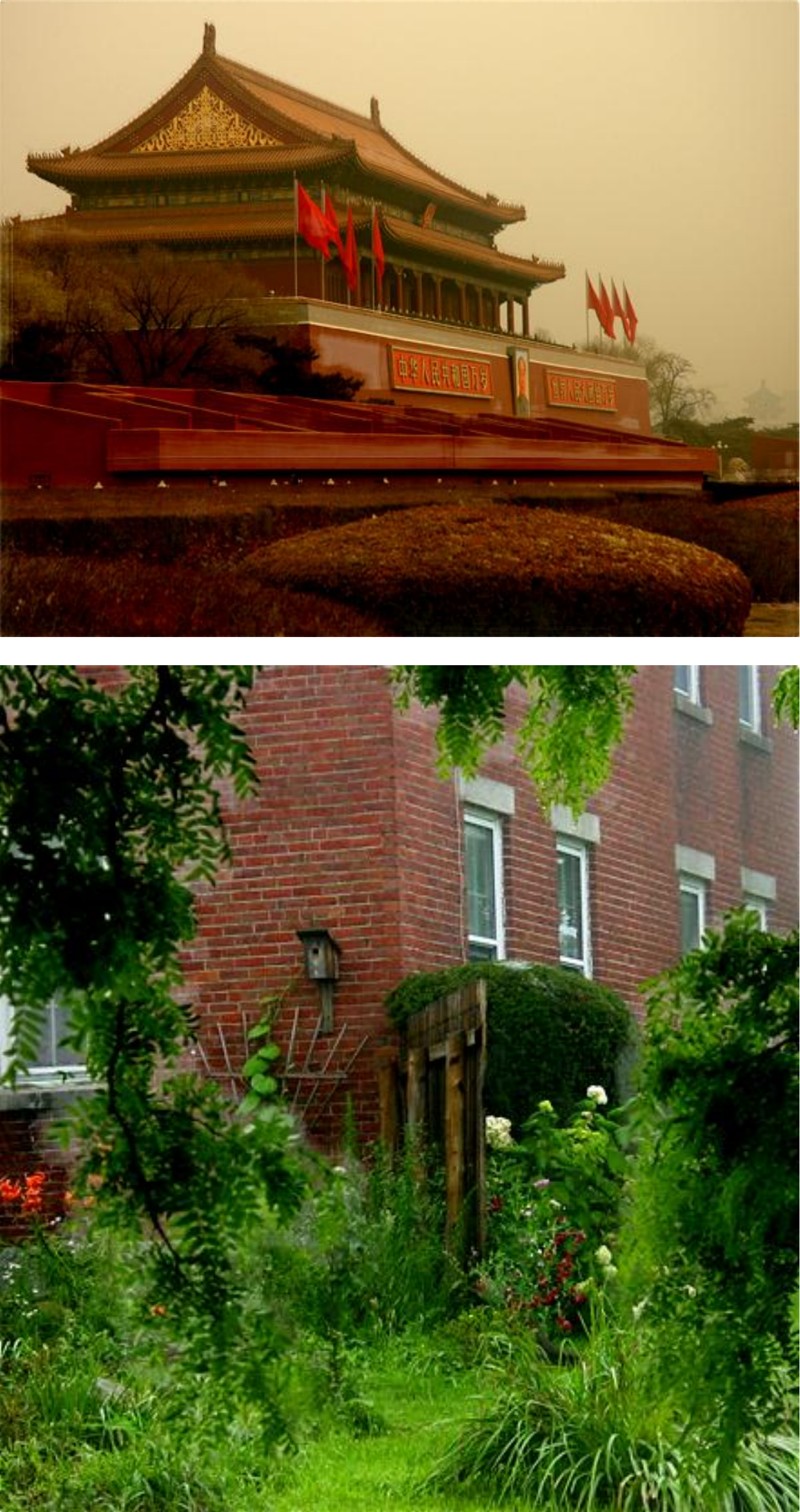}}
  \centerline{(c)}\medskip
\end{minipage}
 \begin{minipage}[b]{0.18\linewidth}
  \centering
  \centerline{\includegraphics[width=2.95cm,height=5.7cm]{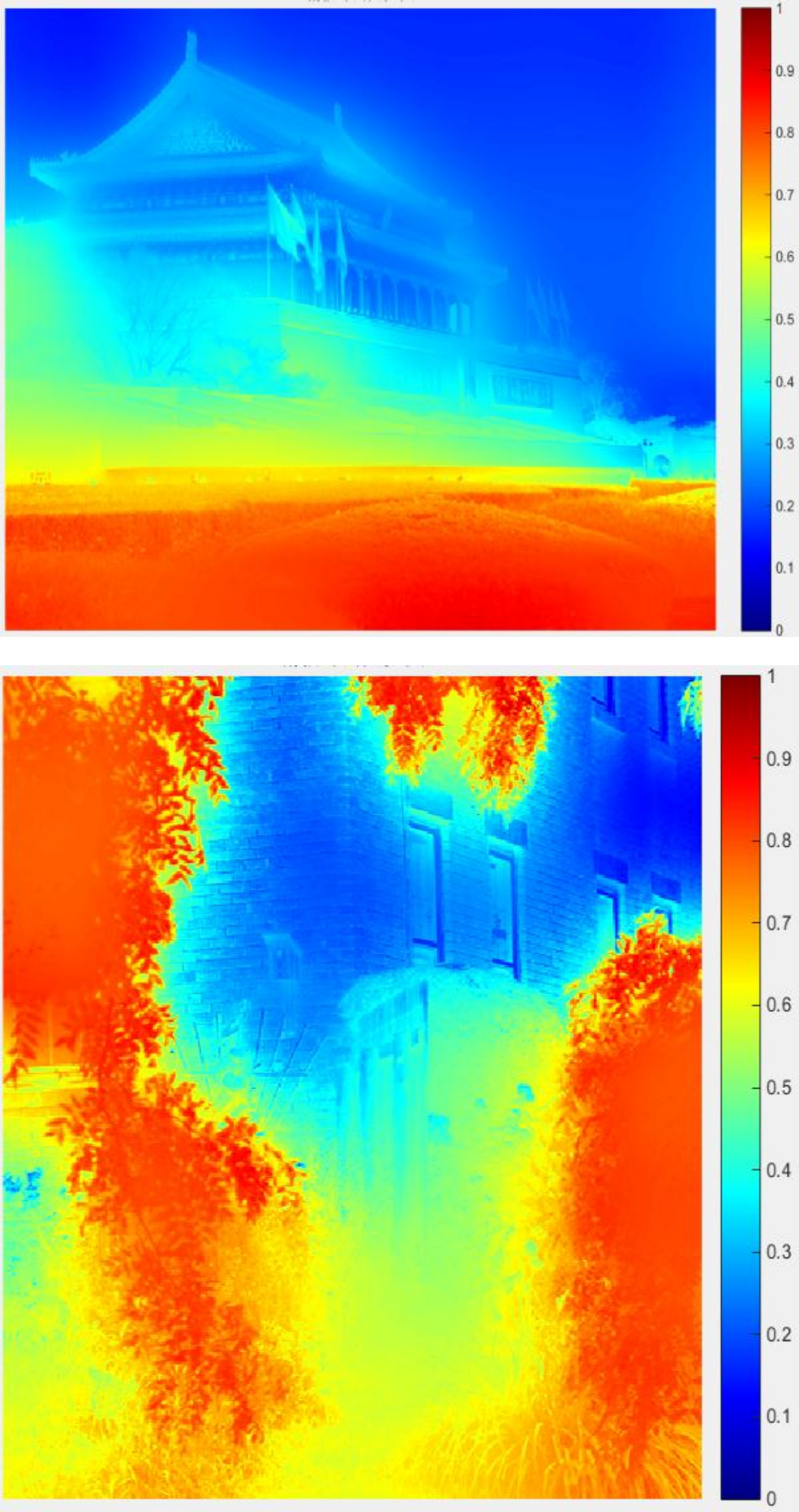}}
  \centerline{(d)}\medskip
\end{minipage}
\begin{minipage}[b]{0.18\linewidth}
  \centering
  \centerline{\includegraphics[width=2.95cm,height=5.7cm]{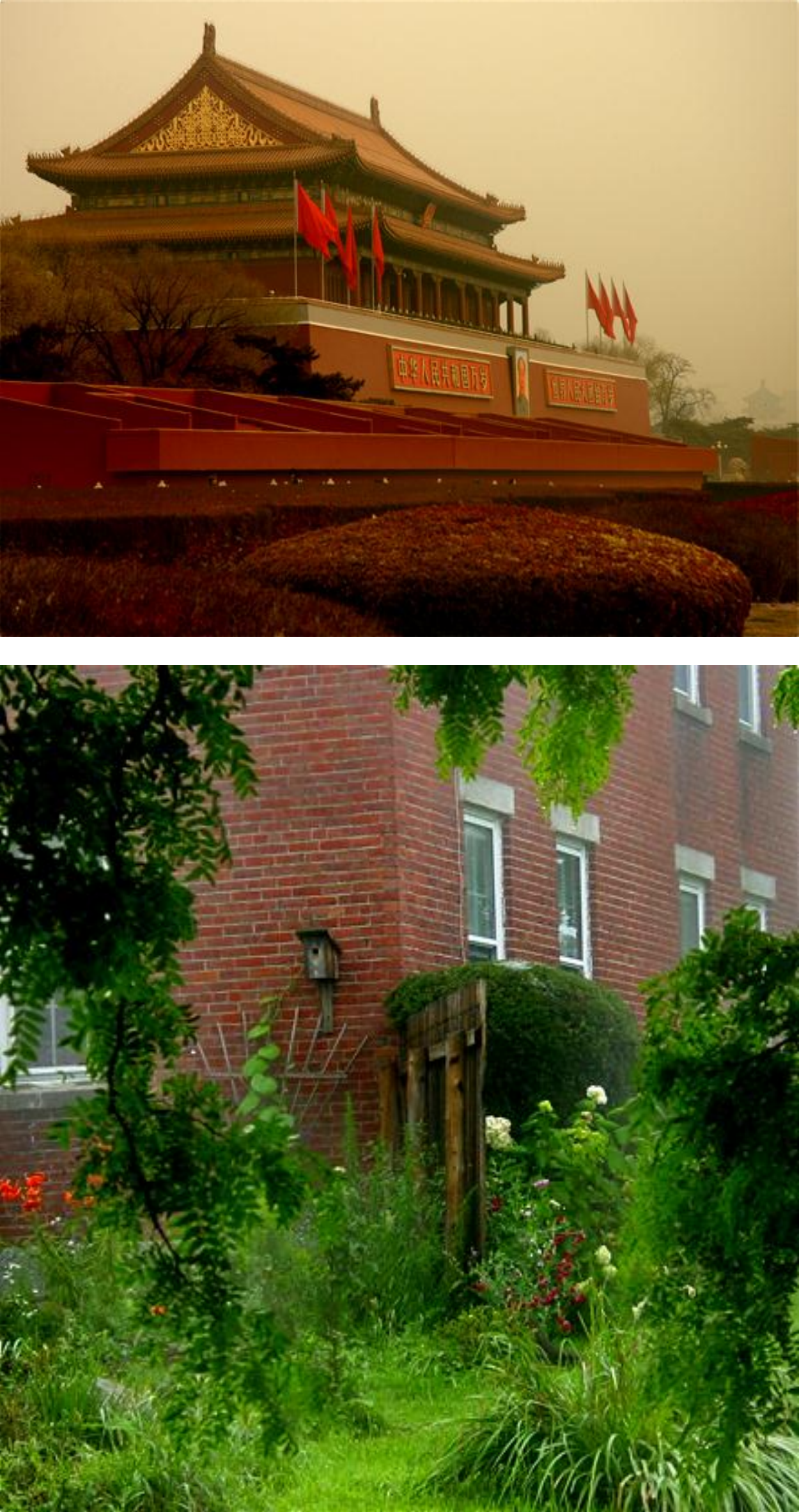}}
   \centerline{(e)}\medskip
\end{minipage}
\caption{Examples of our results. (a) Raw hazy images. (b) The medium transmission estimated by our cascaded network. (c) The dehazed results achieved by our method. (d) The medium transmission estimated by our cascaded network and refined by the guided image filtering \cite{He2013}. (e) The dehazed results achieved by our method using the refined medium transmission. In the medium transmission, different color represents different values (red is close to 1 and blue is close to 0). (Best viewed on high-resolution display with zoom-in.)}
\label{fig:8}
\end{figure*}

In most of patch-based image dehazing methods, after estimating the coarse medium transmission, soft matting \cite{Levin2006} or guided image filtering \cite{He2013} is used to suppress the blocking artifacts. Different from these methods, we observed that our results also look pleasing even though we do not use refinement post-processing (\ie, Figure~\ref{fig:8}(c)). This might be because we optimize the proposed cascaded CNN using full-size images, which reduces the effects of blocking artifacts.

In contrast to the coarse medium transmission (\ie, Figure~\ref{fig:8}(b)), the medium transmission refined by the guided image filtering \cite{He2013} is more smooth and unveils more structure information (\ie, Figure~\ref{fig:8}(d)). Compared with the results in Figure~\ref{fig:8}(c), the results in Figure~\ref{fig:8}(e) have better details and do not have artifacts (\eg, the leaves and parterre). The guided image filtering refinement post-processing is beneficial to our final dehazing performance. Thus, our results shown in this paper are achieved using the medium transmission refined by the guided image filtering. Besides, we do not present the estimated global atmospheric light because it is hard to distinguish the estimated global atmospheric light in figure format. Generally, the accuracy of our global atmospheric light estimation reaches around 90\% in spite of using a light-weight CNN architecture, which also indicates that the task of the global atmospheric light estimation for CNNs is easy.

\section{Experiments}

In this section, we first describe the experimental settings. Then, the effects of network parameter settings are investigated. Finally, we compare the proposed method with several the state-of-the-art single image dehazing methods, such as regularization-based method (Meng \etal \cite{Meng2013}), color attenuation prior method (Zhu \etal \cite{Zhu2015}), and recent CNN-based methods (Cai \etal \cite{Cai2016} and Ren \etal \cite{Ren2016}), on the synthetic and real-world hazy images. The results presented in this paper are achieved by the source code provided by authors.

\subsection{Experimental Settings}
\textbf{Dataset}
There is no easy way to have a amount of the labelled data for our network training. In order to train our cascaded CNN, we generate synthetic hazy images using an indoor RGB-D dataset based on Equation~\eqref{equ_1} and Equation~\eqref{equ_2}.

\begin{figure}[htb]
  \centering
\begin{minipage}[b]{0.3\linewidth}
  \centering
  \centerline{\includegraphics[width=2.5cm,height=6.2cm]{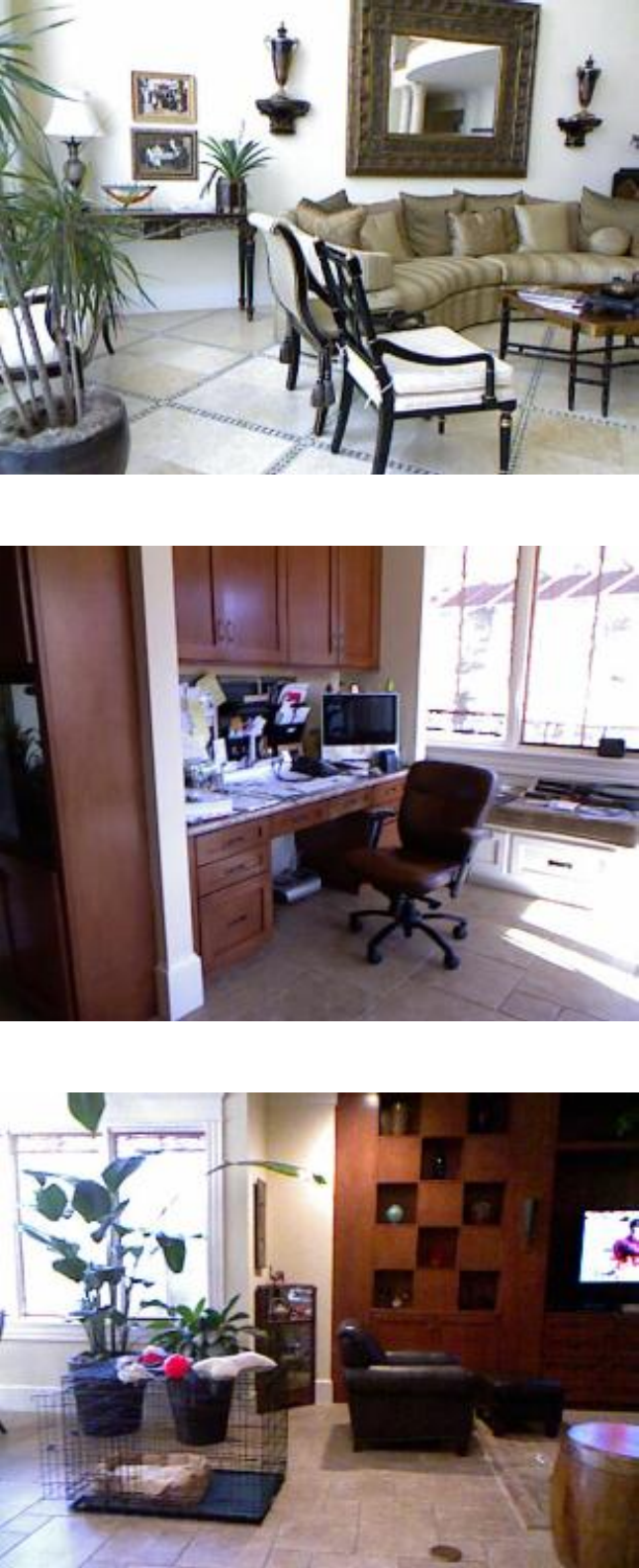}}
  \centerline{(a)}\medskip
\end{minipage}
\begin{minipage}[b]{0.3\linewidth}
  \centering
  \centerline{\includegraphics[width=2.5cm,height=6.2cm]{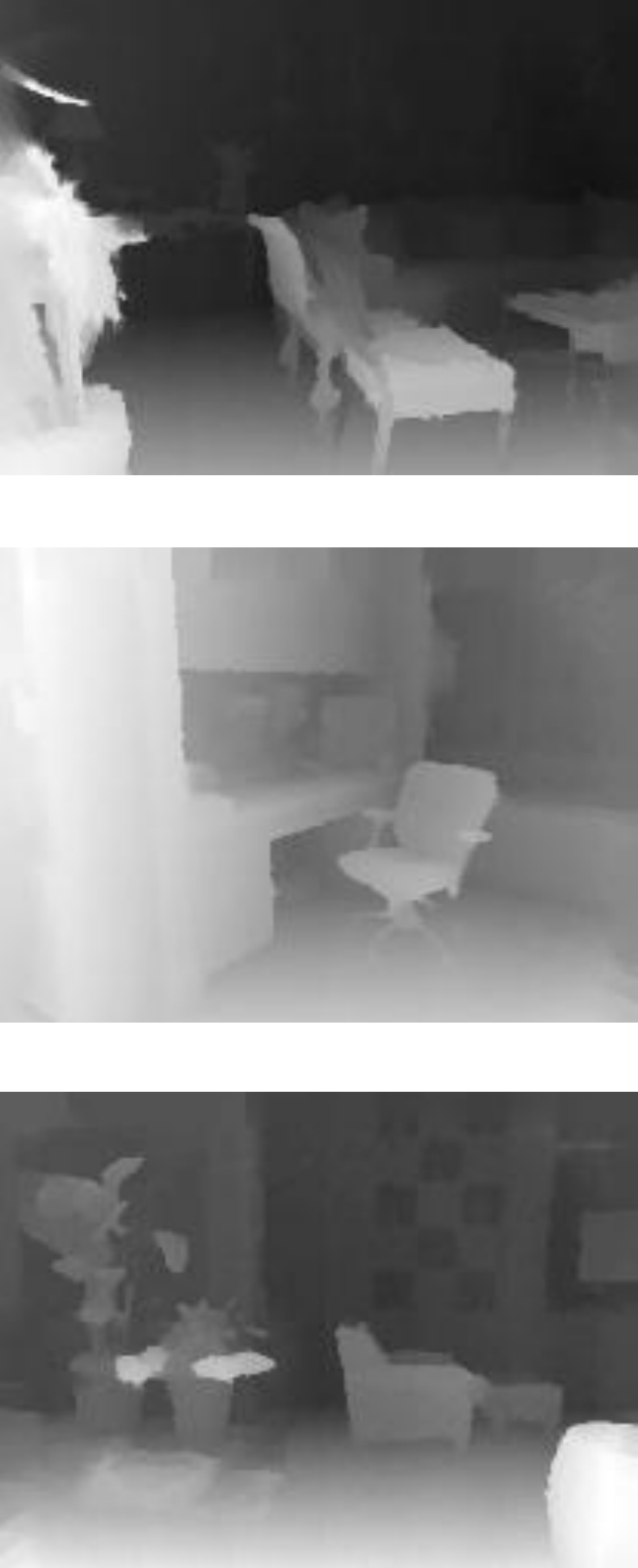}}
  \centerline{(b)}\medskip
\end{minipage}
 \begin{minipage}[b]{0.3\linewidth}
  \centering
  \centerline{\includegraphics[width=2.5cm,height=6.2cm]{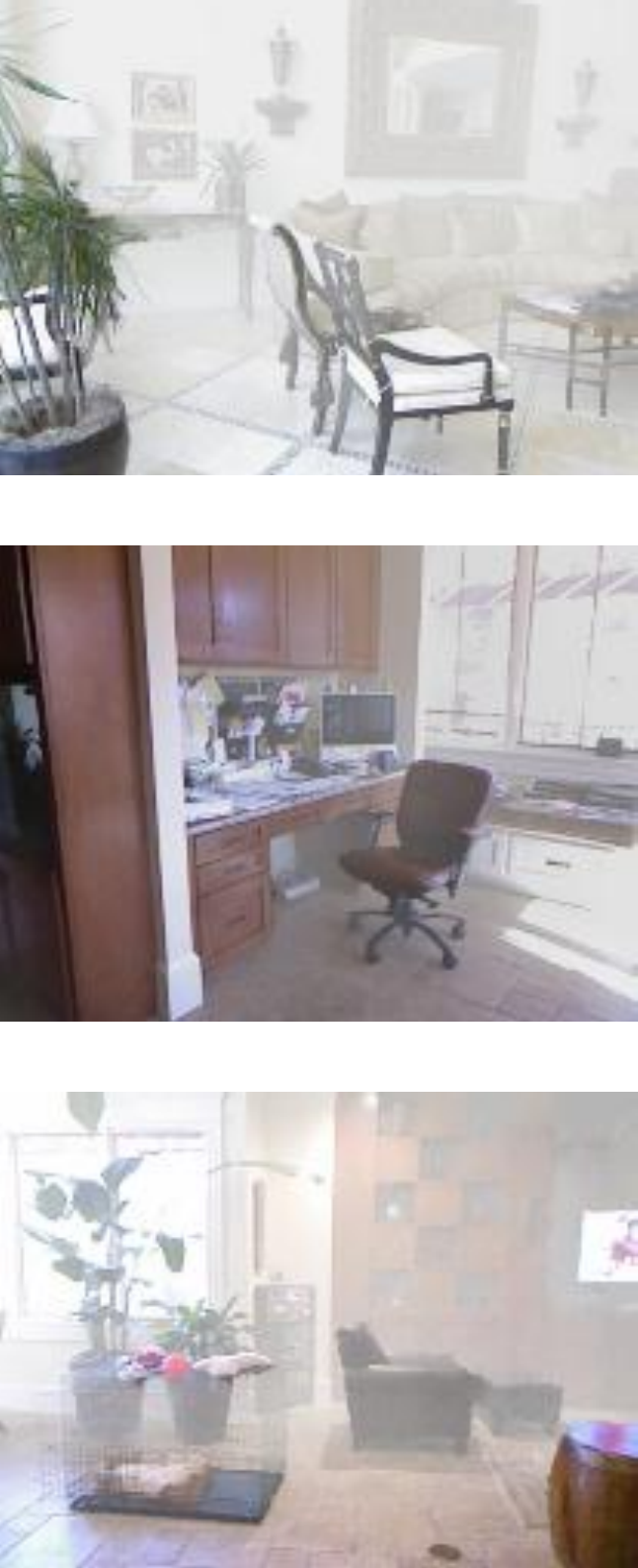}}
  \centerline{(c)}\medskip
\end{minipage}

 \caption{Synthetic samples. (a) Haze-free images from NYU-V2 Depth dataset \cite{Silberman2012}. (b) Synthetic medium transmission maps using the depth images from NYU-V2 Depth dataset \cite{Silberman2012} and the random atmospheric attenuation coefficient based on Equation~\eqref{equ_2}. (c) Synthetic hazy images using (a), (b), and the random global atmospheric light based on Equation~\eqref{equ_1}.}
\label{fig:9}
\end{figure}

Specifically, we assume that (i) the random global atmospheric light $B(x)\in[0.7,1]$; (ii) the atmospheric attenuation coefficient $\beta$ ranging from 0.6 to 2.8 (including haze thickness from light to heavy); (iii) the RGB channels of a hazy image have the same medium transmission and global atmospheric light values. Then, we divide NYU-V2 Depth dataset \cite{Silberman2012} into two parts: one part with 1300 RGB-D images for training data synthesis and another part with 101 RGB-D images for validation data synthesis. For each RGB-D image, we randomly select 5 global atmospheric light and atmospheric attenuation coefficient values to synthesize 5 hazy images. In this way, we synthesize a training set including $1300\times 5$ training samples and a validation set including $101\times 5$ validation samples. Those synthetic samples include hazy images with different haze concentration and light intensities as well as the corresponding medium transmission maps and  global atmospheric light maps. We resize these samples to size $207\times 154$. The depth images in the NYU-V2 Depth dataset have been normalized to [0,1] by us. Figure~\ref{fig:9} presents several synthetic hazy images, the corresponding medium transmission maps, and the haze-free images.

\textbf{Implementation}
In the stage of training our network, the filter weights of each layer are initialized randomly from a Gaussian distribution, and the biases are set to 0. The learning rate is 0.001. The momentum parameter is set to 0.9. A batch-mode learning method with a batch size of 32 is applied. Our cascaded network is implemented by TensorFlow framework. Adam \cite{Kingma2014} is used to optimize our network. The network training with the basic parameter settings shown in Figure~\ref{fig_6} is done on a PC with a Intel(R) i7-6700 CPU @3.40GHz and a Nvidia GTX 1080 Ti GPU.
\subsection{Investigation of Network Parameter Settings}
We mainly investigate the effects of the parameter settings of the shared hidden layers part and the medium transmission estimation subnetwork. Thus, we fix the network parameter settings of the global atmospheric light estimation subnetwork. We do not discuss the parameter settings of the global atmospheric light estimation subnetwork since it already has light enough network weights and reaches high accuracy of the global atmospheric light estimation. The basic filter number of our network is 16, denoted as ${fn}_{16}$, except for the last layer of the medium transmission estimation subnetwork (\ie, 1). The basic filter size used in our network is $3\times 3$, denoted as ${fs}_{3}$. The network depth for the shared hidden layer part and the medium transmission estimation subnetwork is 4 and 7, respectively.

First, we fix other network parameter settings and then only modify one of them. Next, we denote another 3 filter numbers as ${fn}_{8}$, ${fn}_{32}$, and ${fn}_{64}$ and another 2 filter sizes as ${fs}_{5}$ and ${fs}_7$. Besides, we denote another 3 network depth for the shared hidden layer part as ${ds}_{3}$, ${ds}_{7}$ and ${ds}_{9}$. After that, we denote another 2 network depth for the medium transmission estimation subnetwork as ${dm}_{3}$ and ${dm}_{5}$. Here, ${dm}_{3}$ means 3 ``Concat'' blocks in the medium transmission estimation subnetwork. Our basic network architecture includes 2 ``Concat'' blocks. The final loss values on the validation dataset for different network parameter settings are summarized in Table~\ref{table2}. The final loss value for our basic network parameter settings is marked in bold.

\begin{table}[!htbp]
\renewcommand{\arraystretch}{1}
\caption{ The Final Loss for Different Network Parameter Settings} \centering
\begin{tabular}{clcccc}
  \hline
 \textbf{Network Depth} & \textbf{Filter Number} & \textbf{Filter Size}  &\textbf{Loss}\\
 \hline
Basic & Basic & Basic & \textbf{0.043}  \\
Basic & ${fn}_{8}$ & Basic & 0.059 \\
Basic & ${fn}_{32}$ & Basic & 0.033\\
Basic & ${fn}_{64}$ & Basic & 0.018\\
Basic & Basic & ${fs}_{5}$ & 0.030\\
Basic & Basic & ${fs}_{7}$ & 0.028\\
${ds}_{3}$ & Basic & Basic & 0.055\\
${ds}_{7}$ & Basic & Basic & 0.031\\
${ds}_{9}$ & Basic & Basic & 0.023\\
${dm}_{3}$ & Basic & Basic & 0.029\\
${dm}_{5}$ & Basic & Basic & 0.017\\ \hline
\end{tabular}
\vspace{\baselineskip}
 \label{table2}
\end{table}

Finally, we use the above-mentioned basic parameter settings for our cascaded CNN based on its simplicity and efficiency. The cascaded CNN can have varied settings for accuracy and computation trade-off.

\subsection{Comparisons on Synthetic Images}

\begin{figure*}[!htbp]
  \centering
\begin{minipage}[b]{0.13\linewidth}
  \centering
  \centerline{\includegraphics[width=2.3cm,height=13cm]{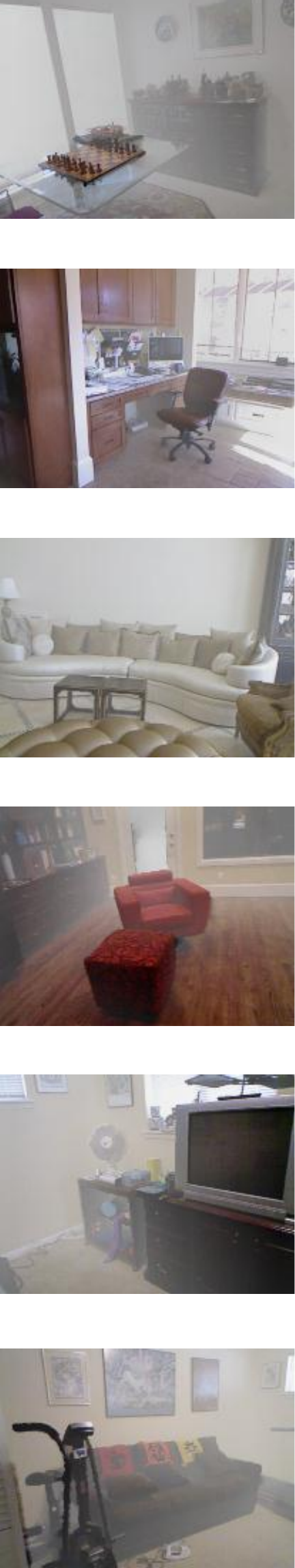}}
  \centerline{(a)}\medskip
\end{minipage}
\begin{minipage}[b]{0.13\linewidth}
  \centering
  \centerline{\includegraphics[width=2.3cm,height=13cm]{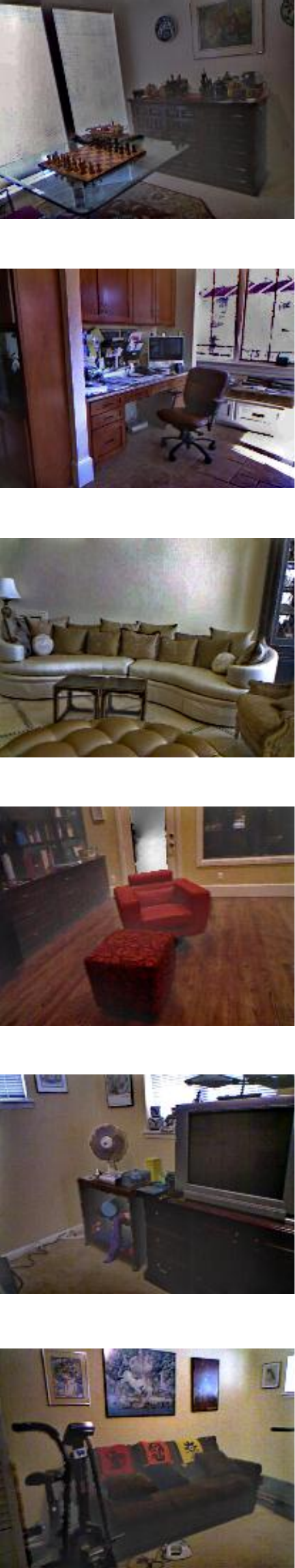}}
  \centerline{(b)}\medskip
\end{minipage}
\begin{minipage}[b]{0.13\linewidth}
  \centering
  \centerline{\includegraphics[width=2.3cm,height=13cm]{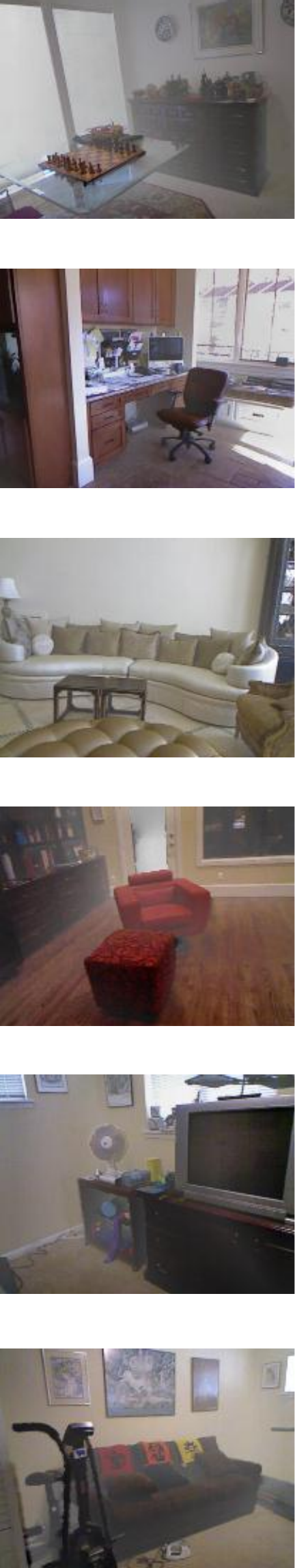}}
  \centerline{(c)}\medskip
\end{minipage}
\begin{minipage}[b]{0.13\linewidth}
  \centering
  \centerline{\includegraphics[width=2.3cm,height=13cm]{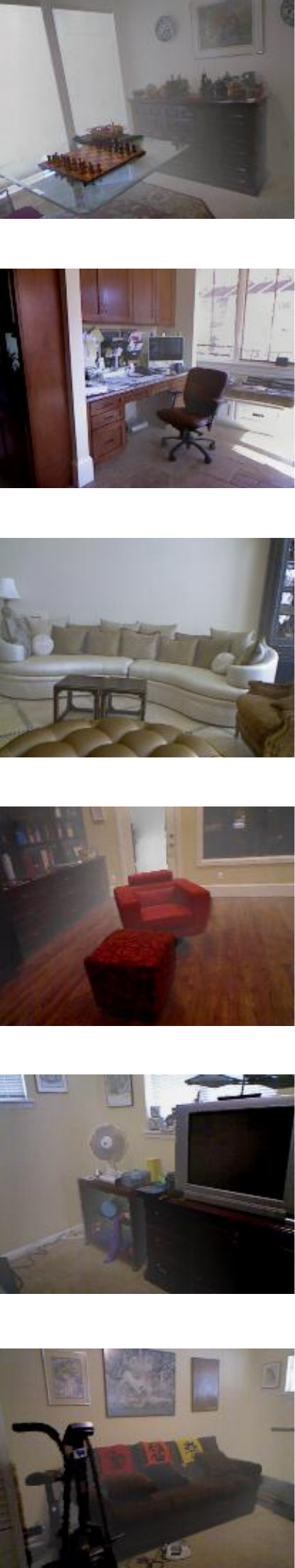}}
  \centerline{(d)}\medskip
\end{minipage}
\begin{minipage}[b]{0.13\linewidth}
  \centering
  \centerline{\includegraphics[width=2.3cm,height=13cm]{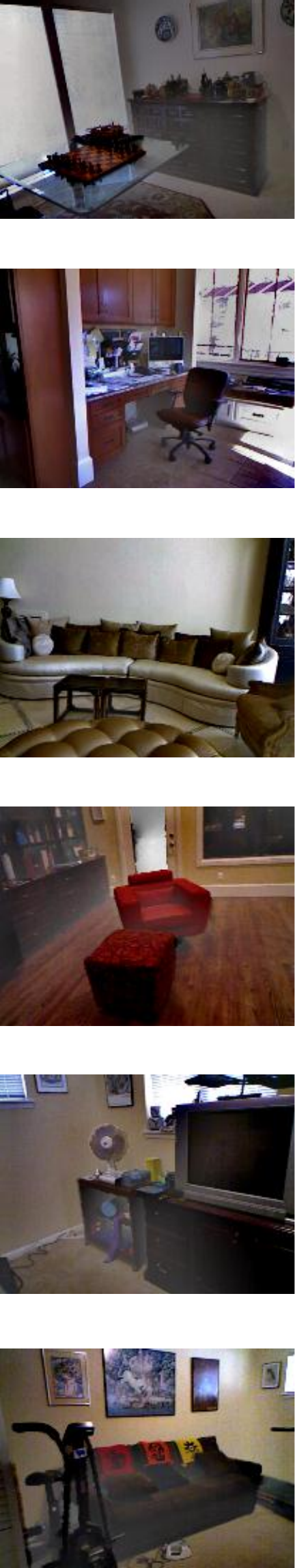}}
  \centerline{(e)}\medskip
\end{minipage}
\begin{minipage}[b]{0.13\linewidth}
  \centering
  \centerline{\includegraphics[width=2.3cm,height=13cm]{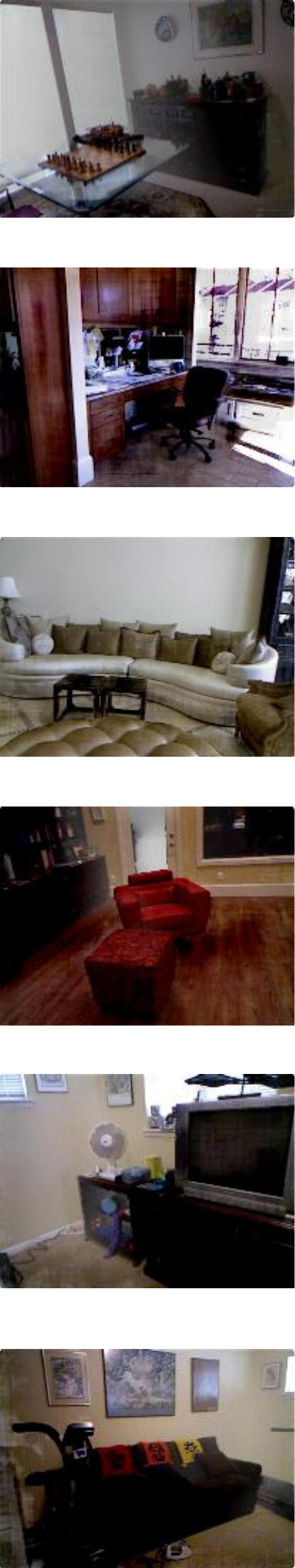}}
  \centerline{(f)}\medskip
\end{minipage}
\begin{minipage}[b]{0.13\linewidth}
  \centering
  \centerline{\includegraphics[width=2.3cm,height=13cm]{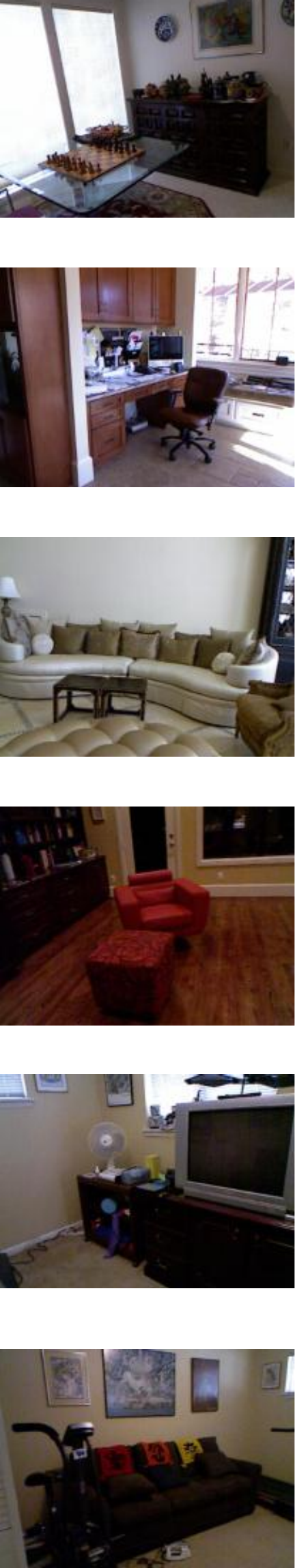}}
  \centerline{(g)}\medskip
\end{minipage}
\caption{ Qualitative comparisons on the synthetic hazy images generated by the same approach with our training data generation. (a) The synthetic hazy images. (b) The results of Meng \etal \cite{Meng2013}. (c) The results of Zhu \etal \cite{Zhu2015}. (d) The results of Cai \etal \cite{Cai2016}. (e) The results of Ren \etal \cite{Ren2016}. (f) Our results. (g) The corresponding haze-free images.}
\label{fig:13}
\end{figure*}

\begin{figure*}[!htbp]
  \centering
\begin{minipage}[b]{0.15\linewidth}
  \centering
  \centerline{\includegraphics[width=2.3cm,height=12cm]{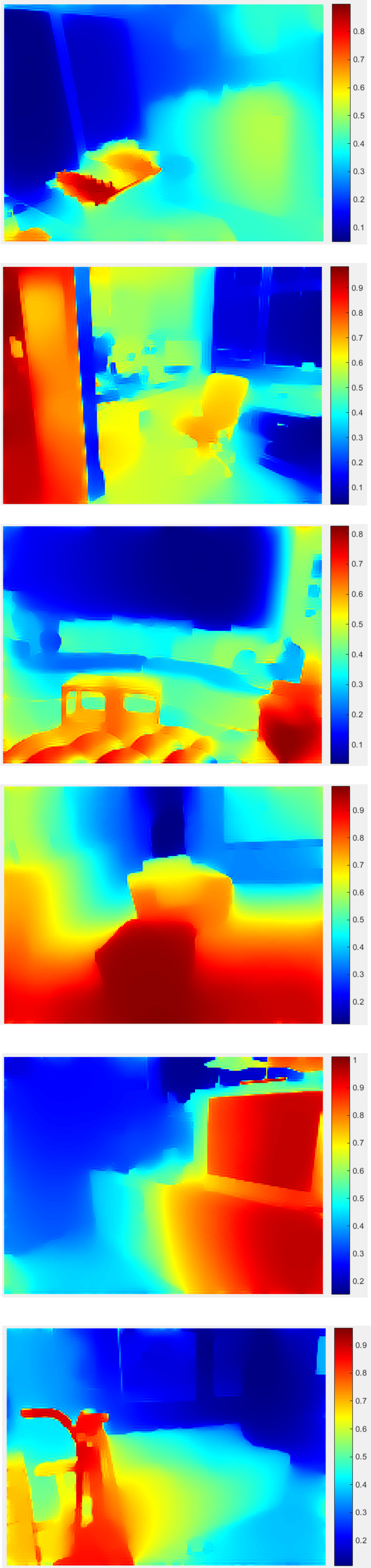}}
  \centerline{(a)}\medskip
\end{minipage}
\begin{minipage}[b]{0.15\linewidth}
  \centering
  \centerline{\includegraphics[width=2.3cm,height=12cm]{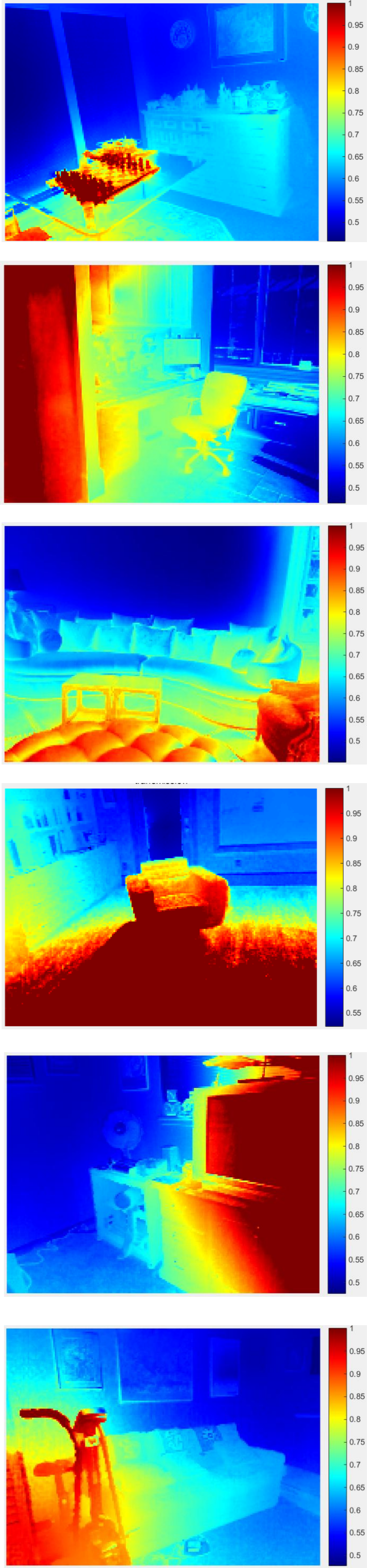}}
  \centerline{(b)}\medskip
\end{minipage}
\begin{minipage}[b]{0.15\linewidth}
  \centering
  \centerline{\includegraphics[width=2.3cm,height=12cm]{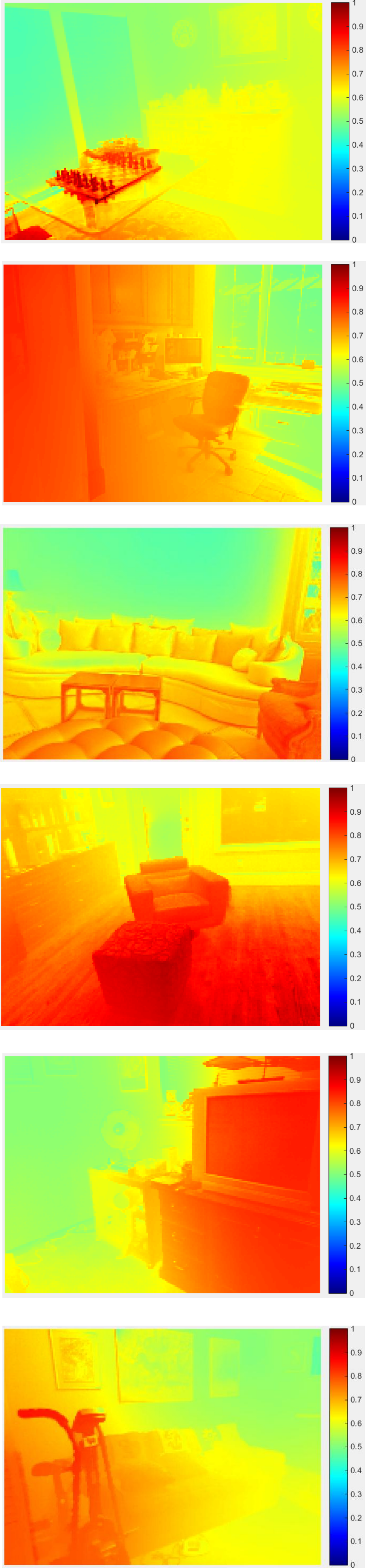}}
  \centerline{(c)}\medskip
\end{minipage}
\begin{minipage}[b]{0.15\linewidth}
  \centering
  \centerline{\includegraphics[width=2.3cm,height=12cm]{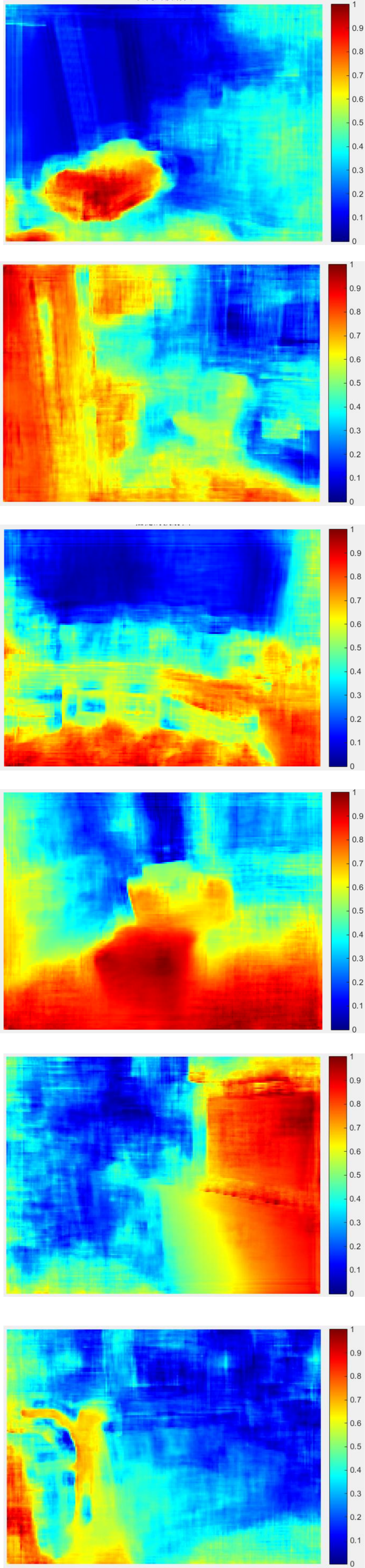}}
  \centerline{(d)}\medskip
\end{minipage}
\begin{minipage}[b]{0.15\linewidth}
  \centering
  \centerline{\includegraphics[width=2.3cm,height=12cm]{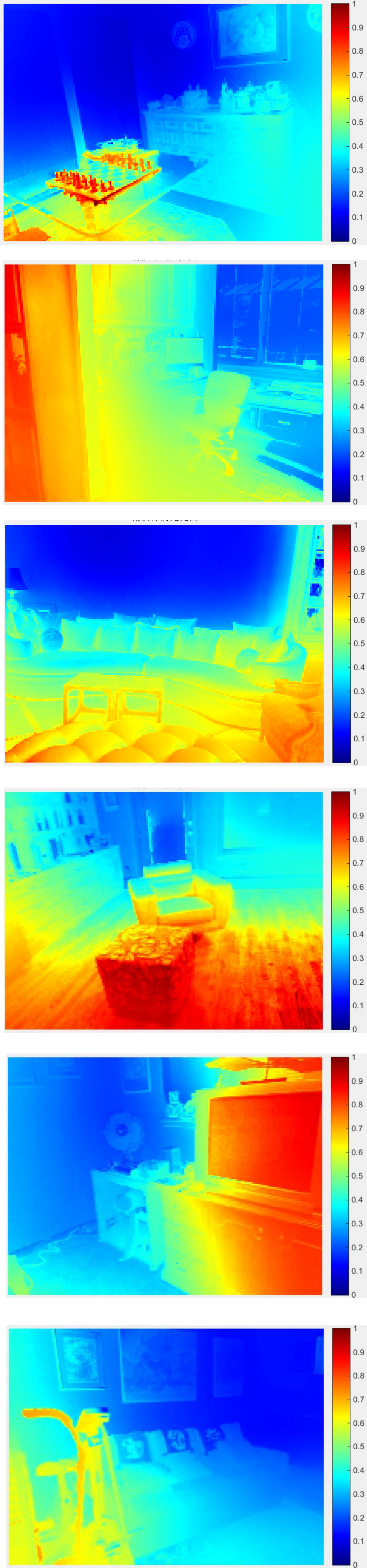}}
  \centerline{(e)}\medskip
\end{minipage}
\begin{minipage}[b]{0.15\linewidth}
  \centering
  \centerline{\includegraphics[width=2.3cm,height=12cm]{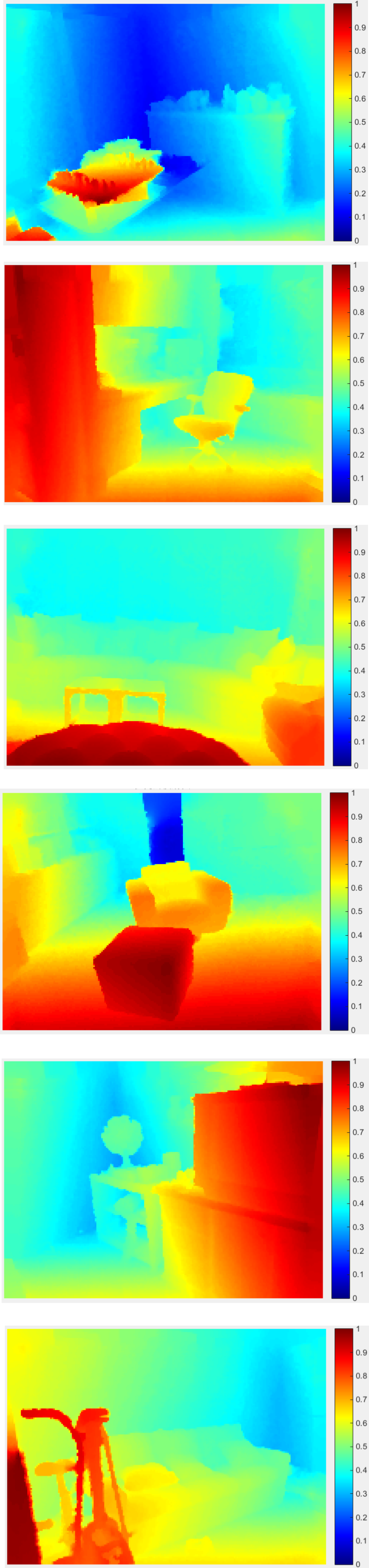}}
  \centerline{(f)}\medskip
\end{minipage}
\caption{ Qualitative comparisons on the estimated medium transmission. (a) The medium transmission estimated by Meng \etal \cite{Meng2013}. (b) The medium transmission estimated by Zhu \etal \cite{Zhu2015}. (c) The medium transmission estimated by Cai \etal \cite{Cai2016}. (d) The medium transmission estimated by our network. (e) The refined results using the guided image filtering \cite{He2013} of (d) . (f) The corresponding medium transmission ground truth.}
\label{fig:133}
\end{figure*}

In this part, we compare our method with the state-of-the-art methods on synthetic hazy images. Firstly, we synthesize a hazy image testing dataset using the same approach with our training data generation. Several dehazed results and the estimated medium transmission on this testing dataset are shown in Figure~\ref{fig:13} and Figure~\ref{fig:133}, respectively.

In Figure~\ref{fig:13}, it is obvious that our method can remove the haze on the input hazy images and restore the color and appearance. Moreover, our results are most close to the ground truth images. It is almost difficult to distinguish our results from the ground truth images. Meng \etal \cite{Meng2013}'s method produces over-enhanced and over-saturated results, while Zhu \etal \cite{Zhu2015} and Cai \etal \cite{Cai2016}'s methods have same dehazing performance which has less effect on the input hazy images, especially for heavy haze. Ren \etal \cite{Ren2016}'s method can remove the haze but still remains haze on several regions. Besides, we also present the corresponding medium transmission estimated by different methods in Figure~\ref{fig:133}. We do not show the medium transmission of Ren \etal \cite{Ren2016} method because the code for medium transmission output is unavailable. In addition, to demonstrate the effectiveness of the refinement post-processing, we also show the coarse medium transmission directly estimated by our medium transmission estimation subnetwork.

As shown in Figure~\ref{fig:133},  all of the estimated medium transmission can indicate the concentration of haze in the hazy images. However, observing Figure~\ref{fig:13}, the dehazed results are different, which indicates that the accuracy of the global atmospheric light estimation is also significant for image dehazing. In addition, compared with the our coarse medium transmission, the refined medium transmission is more smooth, which leads to superior details and textures of our final dehazed results.

Furthermore, we apply the metrics of MSE, Peak Signal-to-Noise Ratio (PSNR) and SSIM \cite {Wang2004} to quantitatively evaluate different methods. The lowest MSE (highest PSNR) indicates that the result is most closed to the corresponding haze-free image in term of image content. The highest SSIM indicates that the result is most close to the corresponding haze-free image in term of image structure and texture. Besides, we also compare the running time (RT) for different methods. The compared methods are implemented in MATLAB and evaluated on the same machine with our model training. Our method is implemented in Python and our RT is calculated on the same machine with the compared methods but with GPU acceleration. Quantitative comparisons are conducted on 50 synthetic hazy images which are generated by the same approach with our training data generation. Some of hazy images and the restored results have been shown in Figure~\ref{fig:13}. Table~\ref{table1} summarizes the average values of the MSE, PSNR, SSIM, and RT for the image with average size of $640\times480$. The values in bold represent the best results.

\begin{table}[!htbp]
\renewcommand{\arraystretch}{1}
\caption{ Quantitative Results on Synthetic Hazy Images in Terms of MSE, PSNR, SSIM, and RT.} \centering
\begin{tabular}{clcccccc}
  \hline
 \textbf{Method} & \textbf{MSE} & \textbf{PSNR (dB)} &  \textbf{SSIM} &\textbf{RT(s)}\\
 \hline
Meng \etal \cite{Meng2013}'s & 7.5742$\ast10^{3}$ & 9.2841& 0.7942 &  2.5460\\
Zhu \etal \cite{Zhu2015}'s  & 6.5163$\ast10^{3}$ & 9.9908 & 0.8355& 0.9486\\
Cai \etal \cite{Cai2016}'s     &  5.1967$\ast10^{3}$ & 10.9635& 0.8474& 1.6905\\
Ren \etal \cite{Ren2016}'s  & 1.2533$\ast10^{3}$ & 17.2971 & 0.8191& 1.8785\\
Ours without refinement     &  1.0991$\ast10^{3}$ &  17.7249 & 0.8607 & 0.0936\\
Ours     &  \textbf{958.1711} &  \textbf{18.3298} & \textbf{0.8857} & \textbf{0.1029}\\
 \hline
\end{tabular}
\vspace{\baselineskip}
 \label{table1}
\end{table}

As shown in Table~\ref{table1}, our method outperforms the compared methods in terms of the average values of MSE, PSNR, SSIM and RT. Moreover, our method without refinement post-processing ranks second, which indicates that refinement post-processing is beneficial to final dehazing performance. Besides, the speed of our method is faster than other methods because of GPU acceleration and our light-weight network parameter settings.
\subsection{Comparisons on Real-World Images}
\begin{figure*}[!htbp]
  \centering
\begin{minipage}[b]{0.15\linewidth}
  \centering
  \centerline{\includegraphics[width=2.5cm,height=15cm]{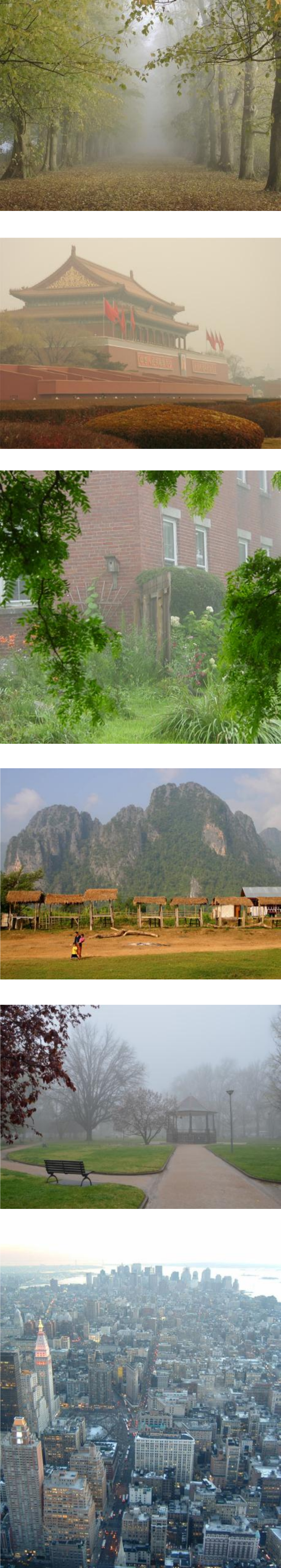}}
  \centerline{(a)}\medskip
\end{minipage}
\begin{minipage}[b]{0.15\linewidth}
  \centering
  \centerline{\includegraphics[width=2.5cm,height=15cm]{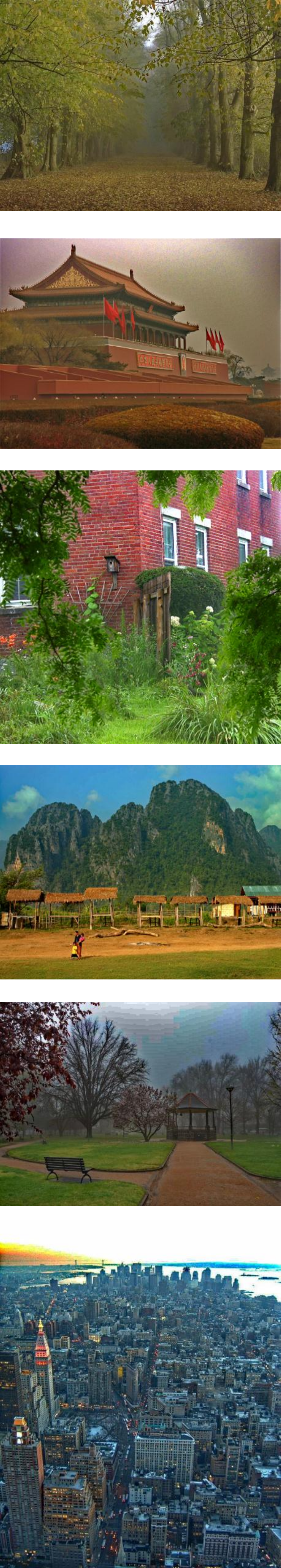}}
  \centerline{(b)}\medskip
\end{minipage}
\begin{minipage}[b]{0.15\linewidth}
  \centering
  \centerline{\includegraphics[width=2.5cm,height=15cm]{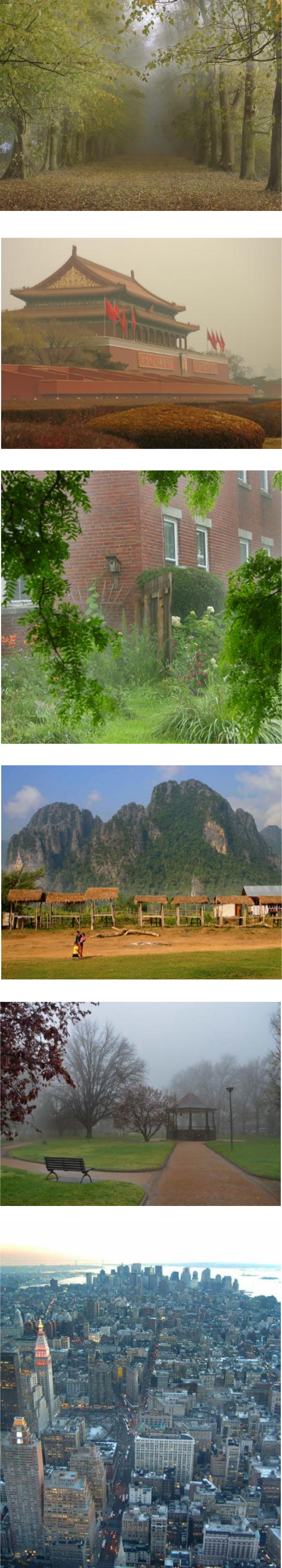}}
  \centerline{(c)}\medskip
\end{minipage}
\begin{minipage}[b]{0.15\linewidth}
  \centering
  \centerline{\includegraphics[width=2.5cm,height=15cm]{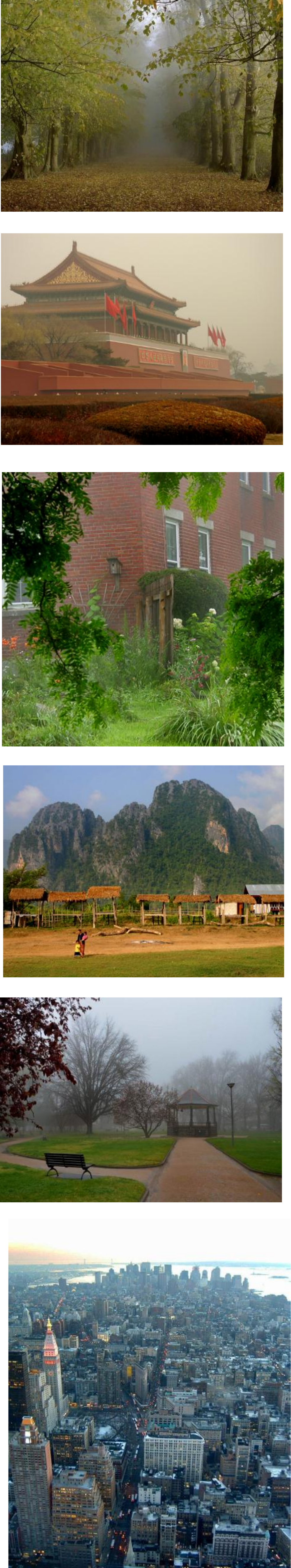}}
  \centerline{(d)}\medskip
\end{minipage}
\begin{minipage}[b]{0.15\linewidth}
  \centering
  \centerline{\includegraphics[width=2.5cm,height=15cm]{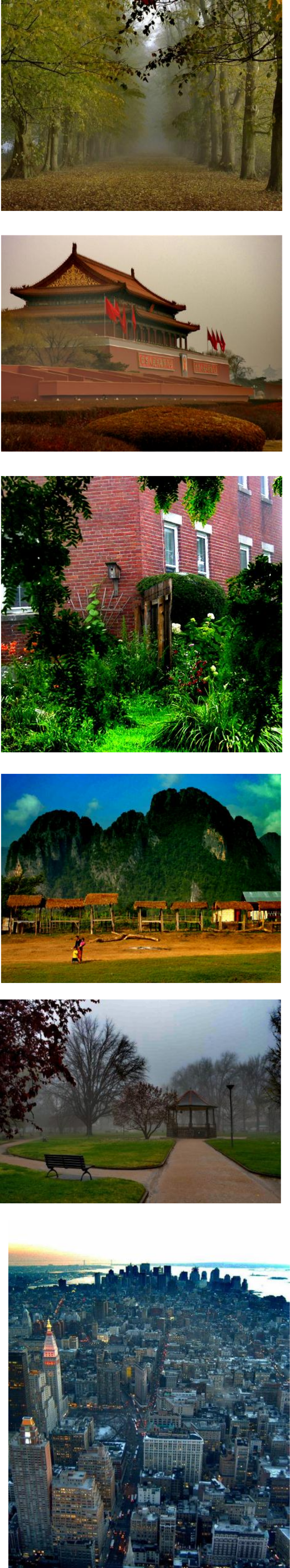}}
  \centerline{(e)}\medskip
\end{minipage}
\begin{minipage}[b]{0.15\linewidth}
  \centering
  \centerline{\includegraphics[width=2.5cm,height=15cm]{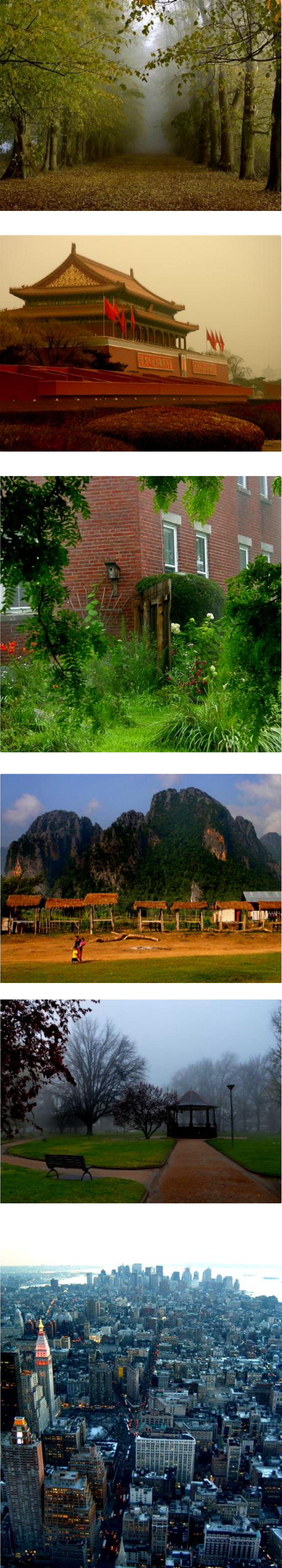}}
  \centerline{(f)}\medskip
\end{minipage}
\caption{Qualitative comparisons on the real-world hazy images. (a) Real-world hazy images. (b) The results of Meng \etal \cite{Meng2013}. (c) The results of Zhu \etal \cite{Zhu2015}. (d) The results of Cai \etal \cite{Cai2016}. (e) The results of Ren \etal \cite{Ren2016}. (f) Our results. (Best viewed on high-resolution display with zoom-in.)}
\label{fig:11}
\end{figure*}

\begin{figure*}[!htbp]
  \centering
\begin{minipage}[b]{0.18\linewidth}
  \centering
  \centerline{\includegraphics[width=2.8cm,height=15cm]{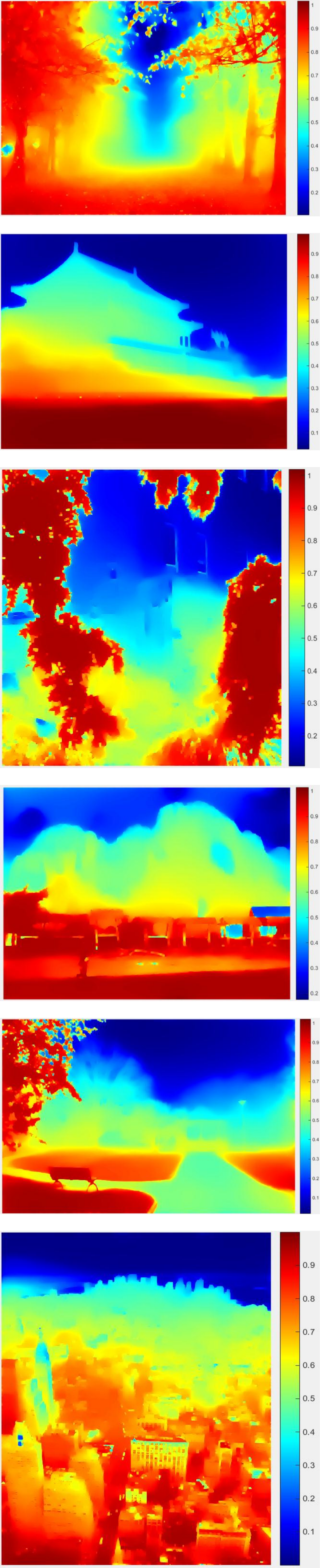}}
  \centerline{(a)}\medskip
\end{minipage}
\begin{minipage}[b]{0.18\linewidth}
  \centering
  \centerline{\includegraphics[width=2.8cm,height=15cm]{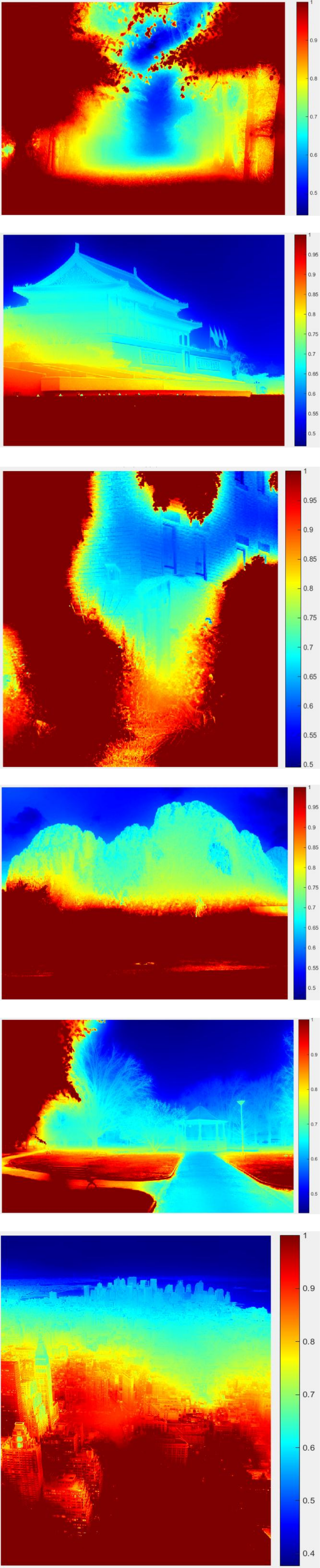}}
  \centerline{(b)}\medskip
\end{minipage}
\begin{minipage}[b]{0.18\linewidth}
  \centering
  \centerline{\includegraphics[width=2.8cm,height=15cm]{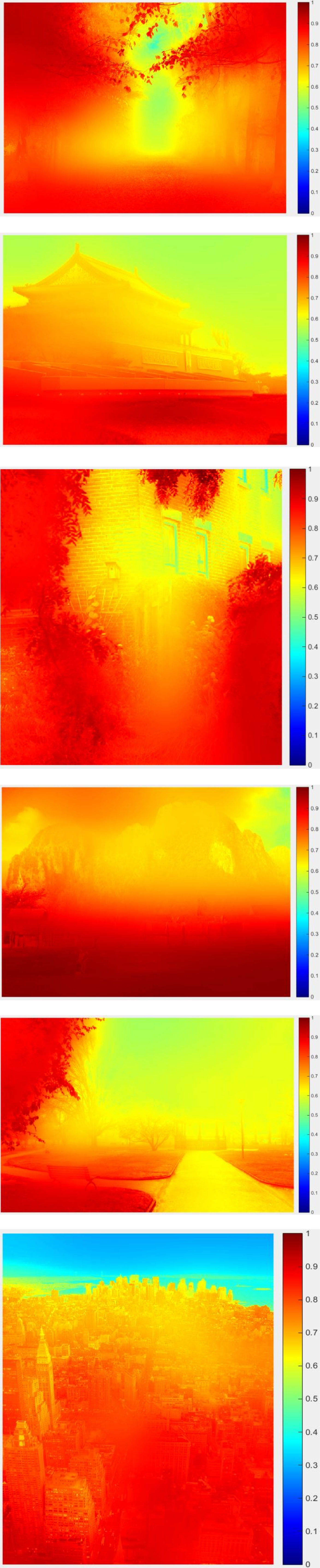}}
  \centerline{(c)}\medskip
\end{minipage}
\begin{minipage}[b]{0.18\linewidth}
  \centering
  \centerline{\includegraphics[width=2.8cm,height=15cm]{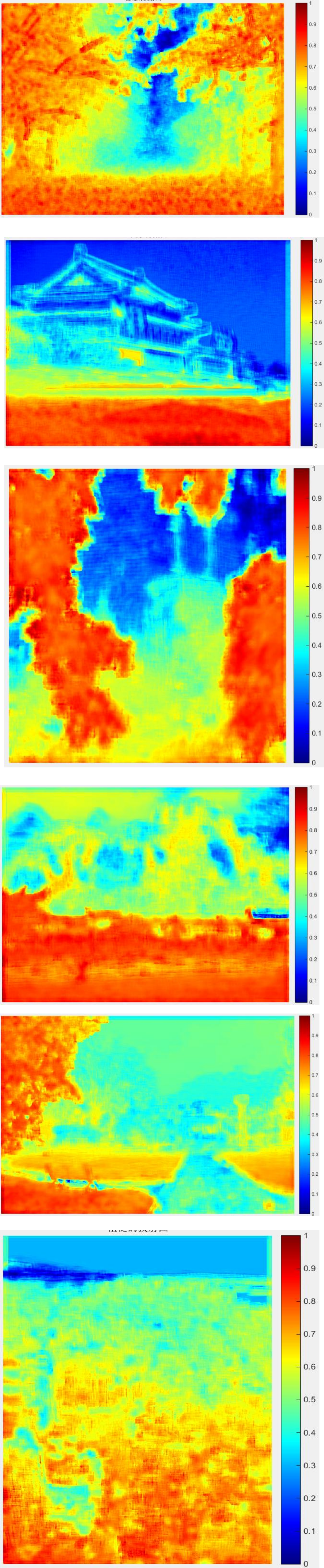}}
  \centerline{(d)}\medskip
\end{minipage}
\begin{minipage}[b]{0.18\linewidth}
  \centering
  \centerline{\includegraphics[width=2.8cm,height=15cm]{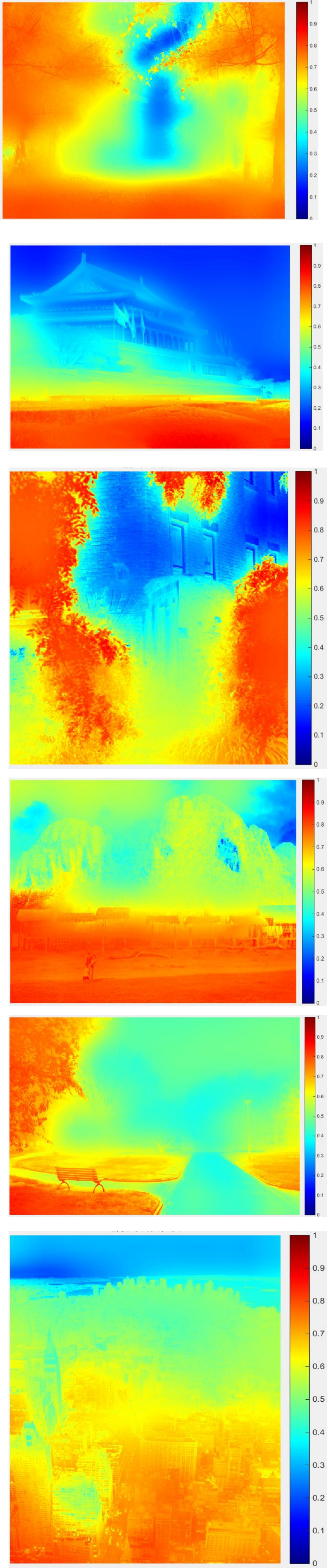}}
  \centerline{(e)}\medskip
\end{minipage}
\caption{The corresponding medium transmission maps of Figure~\ref{fig:11}. (a) The medium transmission estimated by Meng \etal \cite{Meng2013}. (b) The medium transmission estimated by Zhu \etal \cite{Zhu2015}. (c) The medium transmission estimated by Cai \etal \cite{Cai2016}. (d) The medium transmission estimated by our network. (e) The refined results using the guided image filtering \cite{He2013} of (d) . (Best viewed on high-resolution display with zoom-in.)}
\label{fig:12}
\end{figure*}

We conduct several comparisons on the real-world images to verify the performance of the proposed method. We first select several real-world hazy images which are usually used to qualitatively compare and hard to be handled. We compare the proposed method with the above-mentioned state-of-the-art methods in Figure~\ref{fig:11}. Additionally, the corresponding medium transmission is shown in Figure~\ref{fig:12}.

\begin{figure*}[!htb]
  \centering
\begin{minipage}[b]{0.3\linewidth}
  \centering
  \centerline{\includegraphics[width=18.5cm,height=2.1cm]{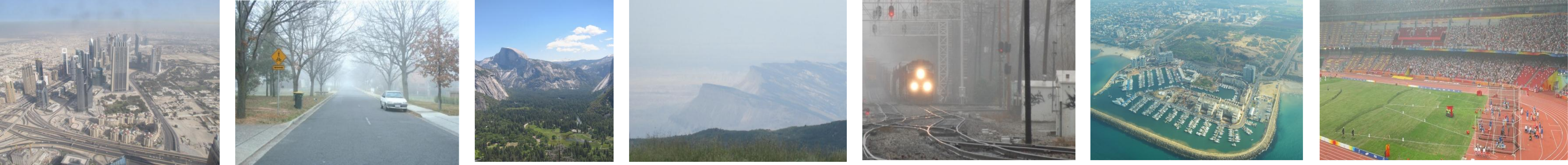}}
  \centerline{(a)}\medskip
\end{minipage}

\begin{minipage}[b]{0.3\linewidth}
  \centering
  \centerline{\includegraphics[width=18.5cm,height=2.1cm]{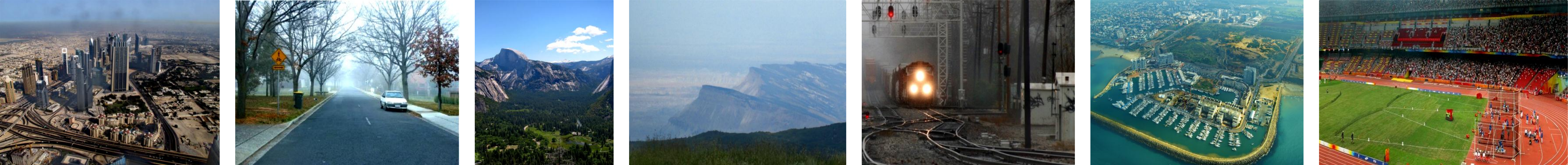}}
  \centerline{(b)}\medskip
\end{minipage}
\caption{Our results on varying scenes. (a) Hazy images. (b) Our results. (Best viewed on high-resolution display with zoom-in.)}
\label{fig:14}
\end{figure*}

In Figure~\ref{fig:11}(b) and Figure~\ref{fig:11}(e), the results of Meng \etal \cite{Meng2013} and Ren \etal \cite{Ren2016} have over-enhanced regions and even introduce color deviation, (\eg, the regions of sky) since these two methods tend to over-estimate the thickness of the haze and are sensitive to sky regions. In Figure~\ref{fig:11}(c) and Figure~\ref{fig:11}(d), the results of Zhu \etal \cite{Zhu2015} and Cai \etal \cite{Cai2016} have significant improvement on the sky regions, but still have some remaining haze on the dense haze regions. Observing Figure~\ref{fig:11}(f), our method produces good dehazing performance in the challenging sky regions and our results have good contrast, vivid color and visually pleasing visibility, which benefits from data-driven non-linear regression. This comparison results are in accordance with those of the synthetic hazy images. Although our cascaded CNN is trained on synthetic hazy images, the experimental results show that our method can be applied for real-world hazy images as well. To further illustrate the performance of different methods, we also present the corresponding medium transmission estimated by the above-mentioned methods in Figure~\ref{fig:12}. Observing Figure~\ref{fig:12}, all of the estimated medium transmission indicates the concentration of haze in the input hazy images. However, the final dehazed results are different, which demonstrates that the global atmospheric light as key component also has significant effects on final results, even for real-world hazy images. Thus, our good dehazing performance benefits from the joint estimation of the global atmospheric light and medium transmission.  More results of our method on challenging hazy images are presented in Figure~\ref{fig:14}.

\section{Discussion and Conclution}

In this paper, we have introduced a novel CNN model for single image dehazing. Inspired by the advances on big data driven low-level vision problems, we formulate a cascaded CNN with special design, which estimates the medium transmission and global atmospheric light jointly. Experimental results show the proposed method outperforms the state-of-the-art methods both on the synthetic and real-world hazy images.

Regarding our method, the remaining question is that similar with most existing image dehazing methods, our method tends to amplify existing image artifacts and noise because our training dataset is generated based on the atmospheric scattering model which does not take artifacts and noise into account. For future work, we intend to suppress artifacts as an integral part in the proposed dehazing model. Additionally, we will investigate end-to-end networks for image dehazing where the networks directly produce haze-free results.

%

\ifCLASSOPTIONcaptionsoff
  \newpage
\fi

\end{document}